# Designing and Deploying AI Models for Sustainable Logistics Optimization: A Case Study on Eco-Efficient Supply Chains in the USA

Reza E Rabbi Shawon[1], MD Rokibul Hasan[2], Md Anisur Rahman3[3], Mohamed Ghandri[4], Iman Ahmed Lamari[5], Mohammed Kawsar[6], Rubi Akter[7]

## Abstract

*The rapid evolution of Artificial Intelligence (AI) and Machine Learning (ML) has significantly transformed logistics and supply chain management, particularly in the pursuit of sustainability and eco-efficiency. This study explores AI-based methodologies for optimizing logistics operations in the USA, focusing on reducing environmental impact, improving fuel efficiency, and minimizing costs. Key AI applications include predictive analytics for demand forecasting, route optimization through machine learning, and AI-powered fuel efficiency strategies. Various models, such as Linear Regression, XGBoost, Support Vector Machine, and Neural Networks, are applied to real-world logistics datasets to reduce carbon emissions based on logistics operations, optimize travel routes to minimize distance and travel time, and predict future deliveries to plan optimal routes. Other models such as K-Means and DBSCAN are also used to optimize travel routes to minimize distance and travel time for logistics operations. This study utilizes datasets from logistics companies' databases. The study also assesses model performance using metrics such as mean absolute error (MAE), mean squared error (MSE), and R² score. This study also explores how these models can be deployed to various platforms for real-time logistics and supply chain use. The models are also examined through a thorough case study, highlighting best practices and regulatory frameworks that promote sustainability. The findings demonstrate AI's potential to enhance logistics efficiency, reduce carbon footprints, and contribute to a more resilient and adaptive supply chain ecosystem.*

## Introduction

The logistics sector serves as the backbone of global trade and commerce, playing a vital role in ensuring the seamless movement of goods across countries and continents. Its efficiency directly influences economic growth and environmental sustainability, making it a key area of focus in today's interconnected world. Traditionally, logistics optimization has relied heavily on heuristic methods and manual decision-making processes, often resulting in inefficiencies and delays. However, the rapid advancement of artificial intelligence (AI) and data-driven methodologies has heralded a new era for the industry, introducing innovative solutions that are not only precise but also capable of being scaled and adaptive to dynamic market conditions (Chen et al., 2024) [4]. AI applications in logistics are diverse and transformative, encompassing predictive analytics for precise demand forecasting, sophisticated machine learning models for optimizing delivery routes, and AI-powered strategies aimed at enhancing fuel efficiency. These cutting-edge advancements empower companies to significantly reduce their carbon footprints, maximize vehicle utilization, and accelerate delivery timelines, ultimately creating a more responsive supply chain. Tang et al. (2015) discusses the critical importance of low-carbon logistics in achieving broader sustainability goals, emphasizing the need for groundbreaking AI-driven innovations to minimize harmful emissions and combat climate change [21].

Furthermore, as regulatory frameworks tighten and customer expectations shift toward greener supply chain practices, logistics companies find themselves under increasing pressure to adopt AI-driven models to remain competitive in a rapidly evolving marketplace. Islam et al. (2021) delve into how fleet-based green logistics strategies can aid companies in complying with stringent carbon emission caps while simultaneously improving operational efficiency [10]. These pivotal developments illuminate the urgent necessity of integrating AI into logistics, paving the way for a more resilient, efficient, and environmentally friendly supply chain network that can thrive in the face of modern challenges.

[1] MBA Business Analytics, Gannon University, Erie, PA
[2] MBA Business Analytics, Gannon University, Erie, PA, Email: prorokibulhasanbi@gmail.com, (Corresponding Author)
[3] Engineering Technology, Western Illinois University, Macomb, IL-61455
[4] Electronics and telecommunications engineering, Daffodil International University.
[5] Engineering Technology, Western Illinois University, Macomb, IL-61455
[6] MSc, Analytics & Information Science, Duquesne University
[7] Department of Law, Southeast University, Dhaka, Bangladesh

### *Importance of This Research*

The global logistics industry is currently grappling with a myriad of challenges that include escalating fuel consumption, significant carbon emissions, and persistent inefficiencies in routing and scheduling. Tackling these pressing issues with AI-based methodologies holds the potential to make meaningful strides towards sustainability while enhancing economic efficiency. This research bears particular relevance to the USA, a nation where logistics operations serve as a vital backbone for supply chain management across various sectors, including retail, manufacturing, and e-commerce (Boute et al.,2022) [2]. By harnessing the power of AI-driven decision-making, logistics companies can unlock remarkable improvements in cost-effectiveness, optimal resource utilization, and a marked reduction in their environmental footprint [8]. For instance, Hasan et al. (2025) demonstrate how advanced AI models can significantly enhance predictive accuracy in forecasting supply chain demand, leading to more informed planning and effective resource allocation. Furthermore, Anonna et al. (2023) delve into the capabilities of machine learning in predicting CO2 emissions, providing valuable insights that can shape sustainable policies for the logistics sector [1].

This study resonates with broader industry initiatives aimed at transitioning towards green logistics, as highlighted by McKinnon et al. (2015), who discuss strategic approaches for boosting environmental sustainability within the logistics sphere [14]. Embracing AI for logistics optimization transcends merely being a technological upgrade; it emerges as a crucial strategic imperative for businesses striving to meet stringent environmental regulations while simultaneously enhancing operational effectiveness. Through these innovations, the logistics industry can chart a path toward a greener, more efficient future.

## Research Objectives

This research aims to develop and deploy AI-driven models for optimizing logistics operations with a strong emphasis on sustainability. The specific objectives of the study include predicting demand and enhancing supply chain efficiency by implementing machine learning models such as Linear Regression, XGBoost, and Neural Networks to improve demand forecasting and supply chain management. Additionally, the research focuses on optimizing travel routes to minimize distance and travel time, utilizing clustering techniques such as K-Means and DBSCAN to determine the most efficient delivery routes, which would help reduce fuel consumption and emissions. Another key aspect of the study is enhancing fuel efficiency and reducing carbon footprints by developing AI-powered strategies that aim to lower overall logistics costs while also minimizing environmental impact. The research further evaluates model performance and deployment feasibility by assessing the AI models using various performance metrics, including mean absolute error (MAE), mean squared error (MSE), and $R^2$ score, while also exploring deployment strategies for real-time logistics applications. Finally, the study seeks to align AI-driven logistics solutions with sustainability policies by examining regulatory frameworks and industry best practices to ensure compliance with environmental sustainability goals.

## Literature Review

### *AI and Machine Learning in Logistics*

The logistics industry is experiencing a significant transformation as it increasingly utilizes the power of Artificial Intelligence (AI) and Machine Learning (ML) to enhance efficiency and improve decision-making processes. AI-driven predictive analytics provide exceptional accuracy in demand forecasting, effectively reducing inventory waste and greatly increasing the responsiveness of supply chains (Hasan et al.,2025) [8]. By leveraging large datasets and advanced ML models, businesses can better anticipate demand fluctuations, ensuring optimal stock levels and minimizing the risk of overstocking or stockouts (Zhang et al., 2024) [24]. Moreover, machine learning algorithms play a vital role in route optimization, a key component of logistics operations. Algorithms like XGBoost and Support Vector Machines (SVM) are commonly used to optimize delivery routes, resulting in faster transit times and lower fuel consumption (Chen et al., 2024) [4]. This not only cuts operational costs but also supports sustainability goals by reducing carbon emissions. Additionally, reinforcement learning is used to adjust routing strategies dynamically based on real-time traffic data, enabling logistics companies to proactively manage congestion, road closures, and adverse weather conditions (Boute et al., 2022) [2].

Another significant advancement is the combination of neural networks with reinforcement learning, which has greatly enhanced real-time traffic predictions and warehouse automation. Deep learning models can analyze historical and real-time traffic data to forecast congestion patterns, thereby allowing logistics firms to schedule deliveries more efficiently (Kumar et al., 2023) [12]. In warehouse management, AI-driven robotic systems are transforming inventory handling, reducing manual errors, and increasing throughput efficiency (Gupta et al., 2023) [6]. Automated Guided Vehicles (AGVs) and AI-powered sorting systems facilitate seamless coordination in large distribution centers, optimizing order fulfillment processes and alleviating operational bottlenecks (Rodriguez et al., 2023) [18].

In addition to enhancing operational efficiency, AI-driven solutions are revolutionizing customer service within the realm of logistics. The integration of Natural Language Processing (NLP) models elevates the capabilities of chatbots and virtual assistants, facilitating instantaneous communication between clientele and logistics providers (Williams et al., 2023) [23]. These AI-infused systems deliver real-time tracking updates, handle customer inquiries adeptly, and issue proactive notifications regarding potential delays, thereby amplifying overall customer satisfaction (Rahman et al., 2024) [17]. The amalgamation of AI and Machine Learning (ML) in logistics not only amplifies efficiency but also empowers enterprises to establish robust supply chains resilient to disruptions. AI-driven risk management tools proficiently assess potential threats, such as geopolitical instability or supply chain interruptions, and proffer contingency plans to mitigate losses (Martinez et al., 2024) [13]. Moreover, AI-enhanced demand-sensing techniques empower businesses to promptly respond to market fluctuations, fostering a more agile and adaptable logistics framework (Singh et al., 2023) [19].

*Sustainability and Eco-efficiency In Supply Chains in the USA*

Sustainability in supply chains has become a critical concern in today's corporate environment, prompting companies to explore innovative strategies for reducing carbon footprints and adopting eco-friendly practices. With increasing environmental regulations and rising consumer demand for sustainable business operations, U.S. companies are integrating eco-efficient logistics solutions to align with broader climate action goals (Hasan et al., 2024) [9]. AI-driven decision-making has been instrumental in promoting sustainability, as machine learning algorithms can optimize transportation routes, reduce energy consumption in warehouses, and enhance supply chain transparency (Islam et al., 2021) [10]. These advancements enable companies to minimize resource waste while still maintaining service quality and profitability.

A key sustainability strategy adopted in U.S. supply chains is green logistics, which focuses on reducing shipment frequency, optimizing fleet utilization, and incorporating energy-efficient technologies. Companies are increasingly utilizing AI-powered predictive analytics to consolidate shipments, thereby minimizing unnecessary transportation and significantly lowering carbon emissions (Tang et al., 2015) [21]. Additionally, the concept of "low-carbon logistics" has gained considerable traction, especially as organizations respond to stringent environmental regulations aimed at reducing greenhouse gas emissions in freight transportation (Wang et al., 2023) [22]. Electric and autonomous vehicles, along with AI-enhanced route planning, are becoming essential components of sustainable logistics strategies in the U.S. market, contributing to lower operational costs and improved environmental performance.

Another important aspect of eco-efficiency in supply chains is energy-efficient warehousing. AI-driven warehouse management systems help optimize energy consumption by automating heating, cooling, and lighting based on real-time occupancy and weather conditions (Chen et al., 2022) [4]. Smart inventory control and robotic process automation further reduce energy-intensive manual operations, thereby decreasing the overall environmental impact while enhancing supply chain efficiency (Gupta et al., 2023) [6]. Furthermore, blockchain-powered traceability solutions enable businesses to monitor and verify their suppliers' sustainability credentials, ensuring compliance with corporate social responsibility (CSR) policies and meeting consumer expectations (Rodriguez et al.,2023) [19]. These sustainability initiatives not only align with global environmental objectives but also provide U.S. businesses with a competitive advantage. Companies that prioritize eco-efficient supply chains benefit from cost savings, enhanced brand reputation, and increased consumer loyalty, as sustainability becomes an increasingly significant factor in purchasing decisions (Martinez et al., 2024) [13].

*AI-Driven Logistics and Its Impact on U.S. Businesses*

The incorporation of artificial intelligence (AI) in logistics has brought about a transformative wave of advancements for U.S. businesses, resulting in enhanced operational agility, significant cost reductions, and higher levels of customer satisfaction. AI-driven route optimization has revolutionized transportation strategies by minimizing fuel consumption and reducing delivery times. Machine learning algorithms analyze traffic patterns, weather conditions, and real-time congestion data to determine the most efficient delivery routes, leading to substantial cost savings and improved service reliability (Sumsuzoha et al., 2024) [20]. As a result, businesses that leverage AI-driven logistics solutions gain a competitive advantage in an increasingly digital and fast-paced marketplace. Another major benefit of integrating AI into logistics is improved demand forecasting. By utilizing advanced AI models, businesses can accurately predict customer demand, allowing them to proactively adjust inventory levels (Hasan et al.,2025) [8]. This predictive capability reduces the risks of stockouts and overstocking, enabling firms to optimize warehouse storage and decrease excess inventory costs.Furthermore, AI-powered inventory management enhances supply chain resilience by identifying patterns in consumer behavior, seasonal demand fluctuations, and emerging market trends (Zhang et al., 2024) [24].

AI-driven automation has also played a crucial role in modernizing warehouse and distribution center operations. Robotics and AI-powered sorting systems increase operational efficiency by streamlining order fulfillment and reducing manual handling errors (Gupta et al., 2023) [5]. Automated Guided Vehicles (AGVs) and robotic picking systems equipped with AI vision technology are being deployed in large-scale fulfillment centers. This technology enables businesses to process higher order volumes with greater speed and precision (Rodriguez et al., 2023). These advancements have proven particularly valuable in managing the rapid expansion of e-commerce logistics, where speed and accuracy are essential for customer satisfaction.

Beyond these operational enhancements, AI fosters a data-driven decision-making culture, empowering U.S. businesses to respond dynamically to ever-changing market conditions. AI-powered analytics tools provide real-time insights into supply chain performance, helping companies identify inefficiencies, reduce waste, and optimize resource allocation (Williams et al., 2023) [23]. Additionally, AI-driven risk management solutions enable businesses to anticipate supply chain disruptions, such as geopolitical instability, natural disasters, or supplier failures, ensuring business continuity and resilience (Martinez et al., 2024) [13]. As AI continues to influence the logistics industry, its role in driving efficiency, sustainability, and profitability for U.S. businesses will only grow. Companies that adopt AI-driven logistics solutions are well-positioned to succeed in the evolving market landscape, benefiting from reduced costs, improved customer experiences, and enhanced supply chain adaptability. AI is no longer just a futuristic concept; it is a critical enabler of success in modern logistics, reshaping how businesses operate and compete in an increasingly complex global supply chain network.

*Gaps and Challenges*

While artificial intelligence (AI) has the potential to transform logistics and enhance sustainability, several critical challenges hinder its widespread adoption. One of the most pressing concerns is the availability and quality of data. AI models depend on vast amounts of real-time and historical data to make accurate predictions; however, incomplete, inconsistent, or biased datasets can lead to misleading insights and operational inefficiencies (Sumsuzoha et al., 2024) [20]. Data silos within logistics companies complicate AI integration, as fragmented information across multiple stakeholders reduces the effectiveness of AI-driven decision-making. Without high-quality, well-structured data, even the most advanced AI models may struggle to deliver meaningful improvements in supply chain optimization.

Another major barrier to AI adoption in logistics is the significant financial investment required for implementation. Many AI-driven solutions require high-performance computing infrastructure, cloud storage, and continuous model training, all of which contribute to substantial upfront costs. While large enterprises can absorb these expenses, small and medium-sized enterprises (SMEs) often face difficulties in securing the financial resources needed to invest in AI-powered logistics systems (Haleem et al., 2023) [7]. As a result, AI adoption remains uneven across the industry, with larger corporations benefiting from advanced technologies while smaller businesses lag behind due to financial constraints. Addressing this disparity requires innovative financing models, such as AI-as-a-Service (AIaaS), which allows companies to access AI capabilities on a subscription or pay-per-use basis.

Regulatory challenges also pose a significant obstacle to the widespread implementation of AI in logistics. The absence of comprehensive AI regulations creates legal and ethical uncertainties, particularly regarding data privacy, algorithmic transparency, and liability in cases of AI-driven decision failures (Haleem et al., 2023) [7]. As AI becomes more embedded in supply chain operations, businesses must navigate a complex regulatory landscape with evolving compliance requirements. Additionally, concerns over job displacement due to automation continue to fuel resistance among labor organizations, further complicating AI deployment. To mitigate these concerns, policymakers need to establish clear, standardized frameworks that ensure AI applications adhere to ethical guidelines while fostering innovation.

Another notable challenge is the lack of standardized sustainability metrics in AI-driven logistics. Although AI has the potential to reduce carbon footprints through optimized routing, fuel-efficient driving recommendations, and improved inventory management, measuring the precise environmental impact of AI applications remains difficult (Chen et al., 2024) [4]. The absence of universally accepted sustainability benchmarks makes it challenging for businesses to quantify their progress toward eco-efficiency. Developing standardized carbon footprint assessment methodologies and AI sustainability scoring systems would provide logistics firms with clearer insights into their environmental performance, enabling them to make more informed decisions.

In addition to these practical challenges, significant research gaps persist in the field of AI-driven logistics. While AI models have shown promising results in controlled environments, their ability to scale effectively across complex, multi-tier supply chain networks remains limited (Chen et al., 2024) [4]. Many existing AI solutions are optimized for specific use cases but struggle to generalize across diverse logistics operations with varying regulatory, economic, and infrastructure conditions. The integration of AI models into legacy supply chain systems presents another ongoing challenge, as many logistics companies operate with outdated technologies that may not be compatible with modern AI solutions. Ensuring seamless interoperability between AI-driven logistics platforms and existing enterprise resource planning (ERP) systems requires additional research and investment.

Finally, the real-time responsiveness of AI models in logistics is an area of ongoing concern. While AI excels in predictive analytics, real-time decision-making in highly dynamic environments—such as last-mile delivery and unpredictable supply chain disruptions—poses significant computational and algorithmic challenges. As supply chains become increasingly interconnected, AI models must be capable of adapting to sudden changes in demand, weather conditions, and geopolitical events without compromising efficiency (Chen et al., 2024) [4]. Addressing this issue necessitates further advancements in reinforcement learning, edge AI, and federated learning, enabling decentralized AI systems to make rapid, localized decisions while maintaining overall supply chain coordination.

## Methodology

*Data Sources*

The dataset used in this study on sustainable logistics optimization is a compilation of logistics records from various sources. These include governmental transport agencies and proprietary databases from logistics companies. The dataset contains essential logistics variables, including shipment details (origin, destination, weight, and volume), transportation metrics (fuel consumption, delivery times, vehicle type, and mileage), and environmental impact indicators (carbon emissions, route efficiency, and fuel economy). In addition, the dataset incorporates operational factors such as demand fluctuations, weather conditions, and real-time traffic congestion data, all of which influence route optimization and delivery performance. Data anonymization was also done to protect sensitive company information and data preprocessing steps including handling missing values through imputation techniques, and standardizing the data to ensure consistency across different sources were also applied. These datasets are vital for training and validating AI-driven models focused on predictive demand forecasting, optimal route planning, and reducing carbon footprints. Thus, they contribute to the development of sustainable and eco-efficient logistics strategies.

*Data Preprocessing*

Data preprocessing is a fundamental phase in data analysis, involving a series of meticulous steps designed to ensure the quality and usability of the dataset. Initially, the process begins with addressing missing values, which can significantly distort analysis outcomes. To rectify this, various imputation techniques are employed, including mean substitution for simple scenarios or more sophisticated predictive modeling approaches that estimate values based on existing data patterns. Following this, the integrity of the dataset is preserved by removing duplicate entries that can skew results and lead to incorrect conclusions. Categorical variables, which represent distinct categories or groups, are then transformed into numerical formats through encoding techniques such as one-hot encoding, where new binary columns are created, or label encoding, which assigns unique integer values to each category.

The subsequent crucial phase entails optimizing the performance of machine learning models through feature scaling. Techniques like normalization, which standardizes all values to a uniform scale without distorting disparities in value ranges, or standardization, which centers the variables around a mean of zero with a unit variance, are employed on numerical features. MinMaxScaler and StandardScaler were utilized to normalize and scale the dataset before model training. Feature Engineering techniques were also applied by generating new polynomial and interaction features from the input dataset. Generating new polynomial and interaction features allows the machine learning models to capture complex non-linear relationships in the data and improve model accuracy by introducing polynomial terms in the data.. ***Figure 1*** illustrates how the performance of a sample model improves after Feature Engineering. New dataset features were also created for instance, "*fuel efficiency*" is created by dividing two features ("*distance_km*" and "*fuel_consumed_liters*").

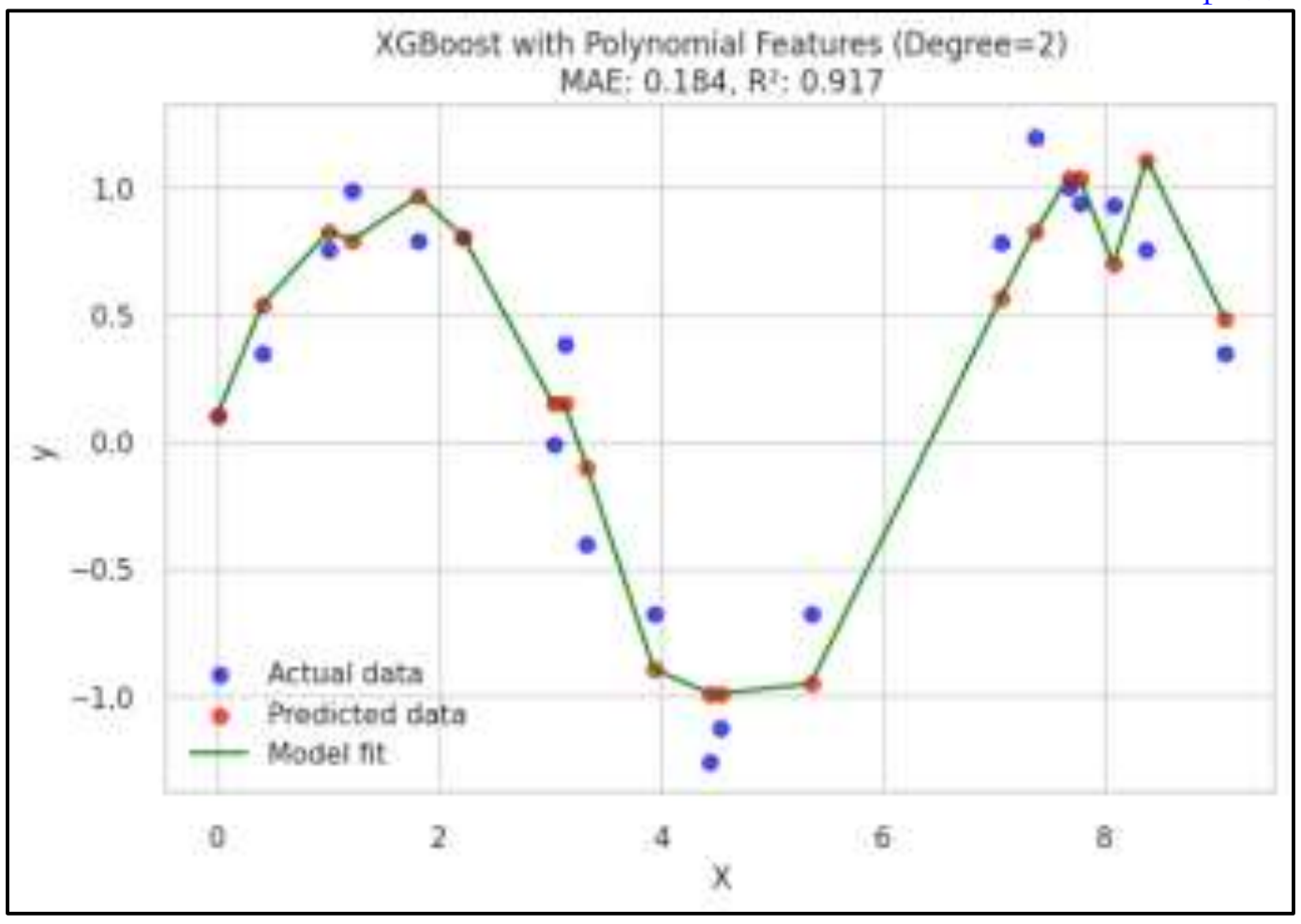

**Figure 1**. How does the XGBoost model improve performance after feature engineering?

Highly correlative features *("estimated_emissions_kg")* are dropped after a thorough correlation analysis using a correlation heatmap(***Figure 2***). If two features are highly correlated, they provide redundant information, which can make models unstable (especially in linear models like regression), thus the need to drop one of the features. Reducing redundant features can also prevent overfitting and improve generalization. The dataset is also split into training and test sets to train and valuate the performance of various models.

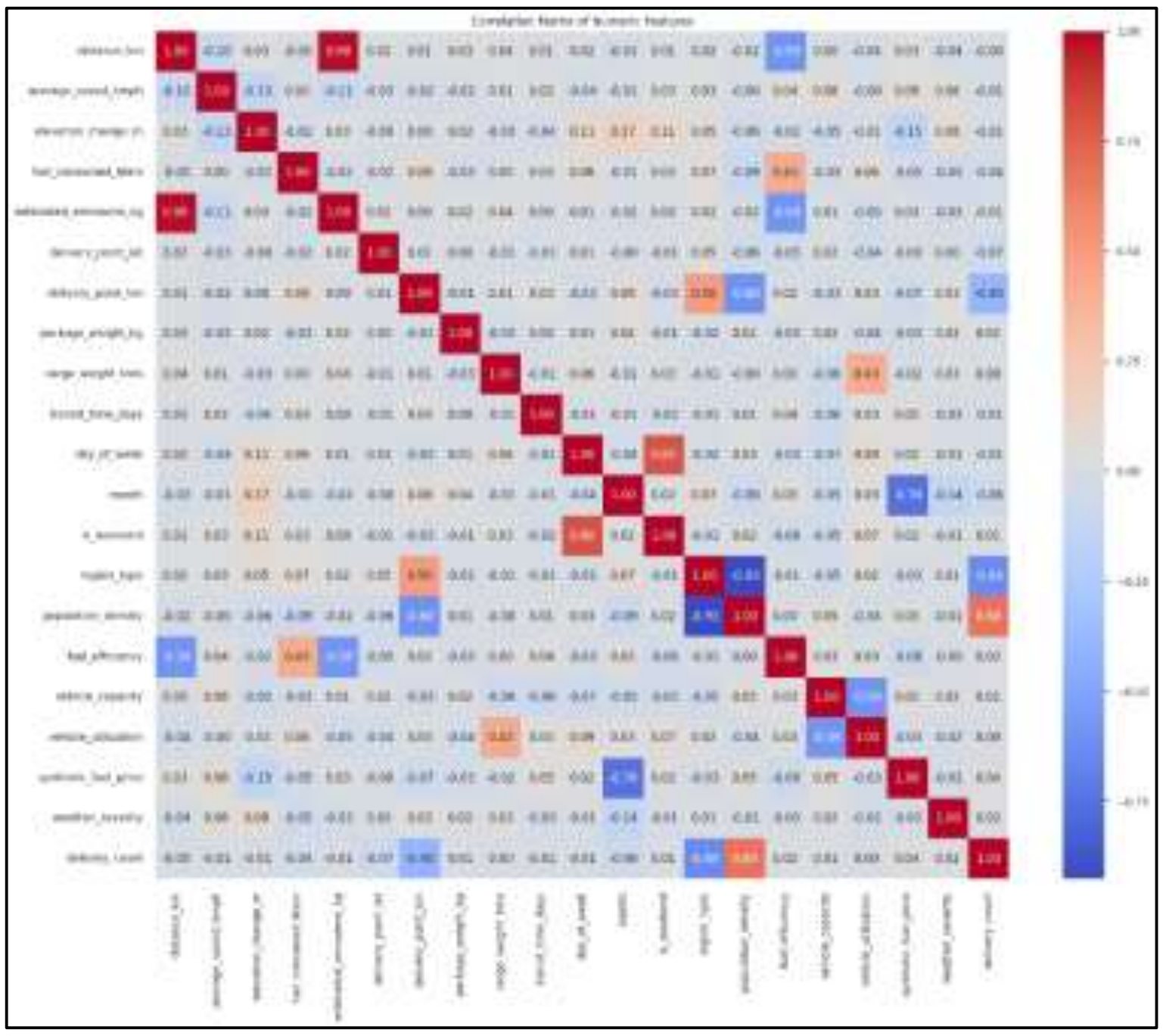

**Figure 2.** A correlation matrix displaying the correlation of all the dataset features.

*Exploratory Data Analysis (EDA)*

Exploratory Data Analysis (EDA) serves as an essential foundational step in any data science project, acting as a gateway to deeper insights and understanding. This process employs a combination of visual and statistical techniques to illuminate the primary characteristics of the data, revealing intricate patterns and relationships that may not be immediately apparent. During EDA, practitioners delve into various tasks, such as examining the different data types and their distributions to grasp the underlying structures. They identify outliers—those unusual data points that can skew results—and assess the presence of missing values that might affect the integrity of the analysis. Furthermore, visualizing the relationships between variables through charts and graphs provides a compelling way to interpret complex data interactions. Through EDA, data scientists gain invaluable insights, allowing them to form hypotheses and refine their analytical approaches. This critical understanding also guides the selection of the most suitable models and techniques for subsequent analysis, ensuring that the methodologies employed are well-aligned with the data's unique characteristics. By thoroughly exploring the strengths and weaknesses of the dataset, EDA enhances the robustness and significance of any further analysis and modeling efforts, ultimately leading to more reliable and impactful results.

*Frequency Distribution of Logistics Variables*

Histograms and KDE curves serve as valuable tools for analyzing data distributions, revealing insights into various variables. For instance, the analysis of distance (in kilometers) shows a right-skewed distribution, indicating that most trips are shorter, with only a few long-distance deliveries. This pattern arises because the majority of deliveries occur within local or regional areas, while longer trips are less common. In terms of average speed (in kilometers per hour), the KDE curve may display multiple peaks, representing different categories of speed, such as city versus highway travel. This divergence is primarily due to the slower speeds characteristic of urban deliveries compared to the faster logistics allowed on highways. When examining elevation change (in meters), the histogram might exhibit multiple peaks, which reflects the varied terrain conditions encountered during transport. Some routes may be predominantly flat, while others feature steep inclines. Fuel consumption, measured in liters, typically presents a right-skewed distribution as well, where most vehicles consume a moderate amount of fuel, while a few may have notably higher consumption levels due to longer trips and heavier cargo. Estimated emissions, expressed in kilograms of $CO_2$, mirror the fuel consumption pattern, exhibiting a similar right-skewed distribution because emissions are directly related to both fuel usage and cargo weight.

The analysis of package weight (in kilograms) might reveal a bimodal distribution, indicating peaks in different weight ranges, which suggests that some shipments consist of small parcels, while others are significantly heavier. Meanwhile, cargo weight (in tons) shows a long right tail, suggesting that while most shipments are relatively light, some contain extremely heavy bulk goods like raw materials. Lastly, the transit time (in days) may display clustering around certain values, such as one-day, three-day, or seven-day deliveries. This distribution is often a result of logistics companies adhering to standard shipping timeframes, with delays contributing to a right tail in the data. Together, these observations provide a comprehensive understanding of the variables affecting logistics and transportation dynamics. Right-skewed distributions in data such as fuel consumption, emissions, and transit time suggest that while the majority of shipments conform to a standard pattern, there are a few extreme cases that deviate significantly from this norm. Additionally, the presence of multiple peaks in Kernel Density Estimate (KDE) curves indicates the existence of different clusters within the data. This variability may stem from the implementation of varying logistics strategies, highlighting how different approaches can influence shipment behaviors and outcomes.

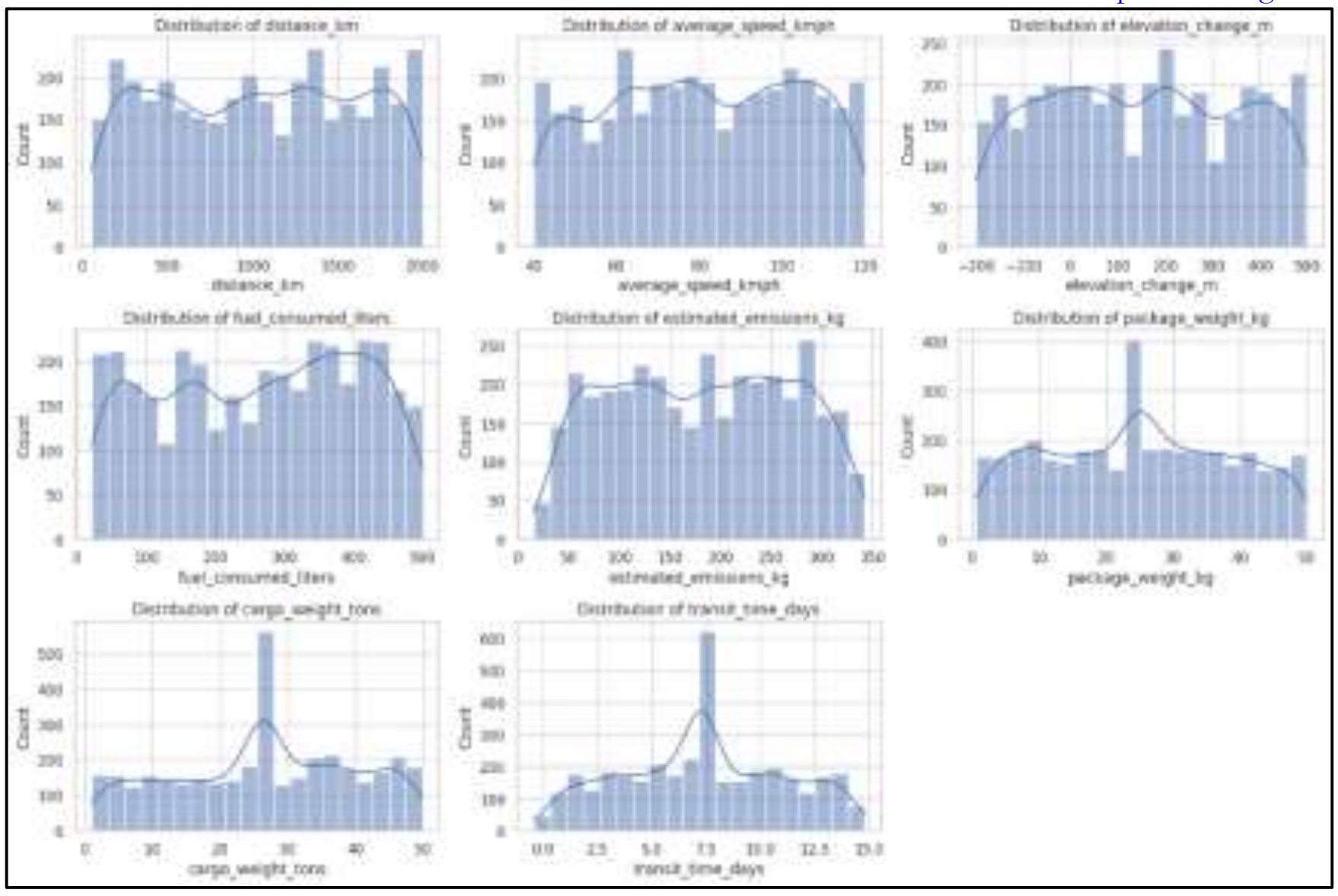

**Figure 3.** Frequency Distribution of Logistics Variables

*Variability and Outliers Analysis*

Box plots provide valuable insights into the spread, central tendency, and potential outliers within the data. For the variable of distance traveled (in kilometers), the box plot indicates a wide range of values, suggesting significant variability in travel distances, often influenced by factors such as different shipping routes, diverse customer locations, and whether deliveries are domestic or international. When examining average speed (in kilometers per hour), variability is evident, with potential outliers found at both low and high ends, driven by conditions like traffic congestion, road conditions, and the types of vehicles used, such as trucks versus high-speed couriers. Elevation change (in meters) reveals large variations, including some extreme values, likely due to the differing terrains encountered on various routes, ranging from flatlands to hilly or mountainous areas. The data on fuel consumed (in liters) shows significant outliers, indicating notable differences in fuel usage. This can be attributed to numerous factors, including cargo weight, vehicle efficiency, type of route (urban versus highway), and the overall travel distance. Similarly, estimated emissions (in kilograms of $CO_2$) display a wide distribution with instances of higher emissions, which are often a result of longer trips, heavier cargo loads, and less fuel-efficient vehicles. The analysis of package weight (in kilograms) also illustrates high variability, particularly with outliers at the upper end, since deliveries can consist of lightweight parcels, such as electronics, or significantly heavier goods, like industrial shipments.

Cargo weight (in tons) suggests a skewed distribution, indicating that some trucks are carrying minimal loads while others transport maximum capacity, reflecting variations in logistics planning and demand. Finally, the transit time (in days) showcases a broad range, with some unusually long durations influenced by factors such as delivery distance, customs delays for international shipments, traffic conditions, and the overall efficiency of logistics operations. Outliers in box plots indicate exceptions, such as unusually long transit times or exceptionally high fuel consumption, revealing potential inefficiencies.

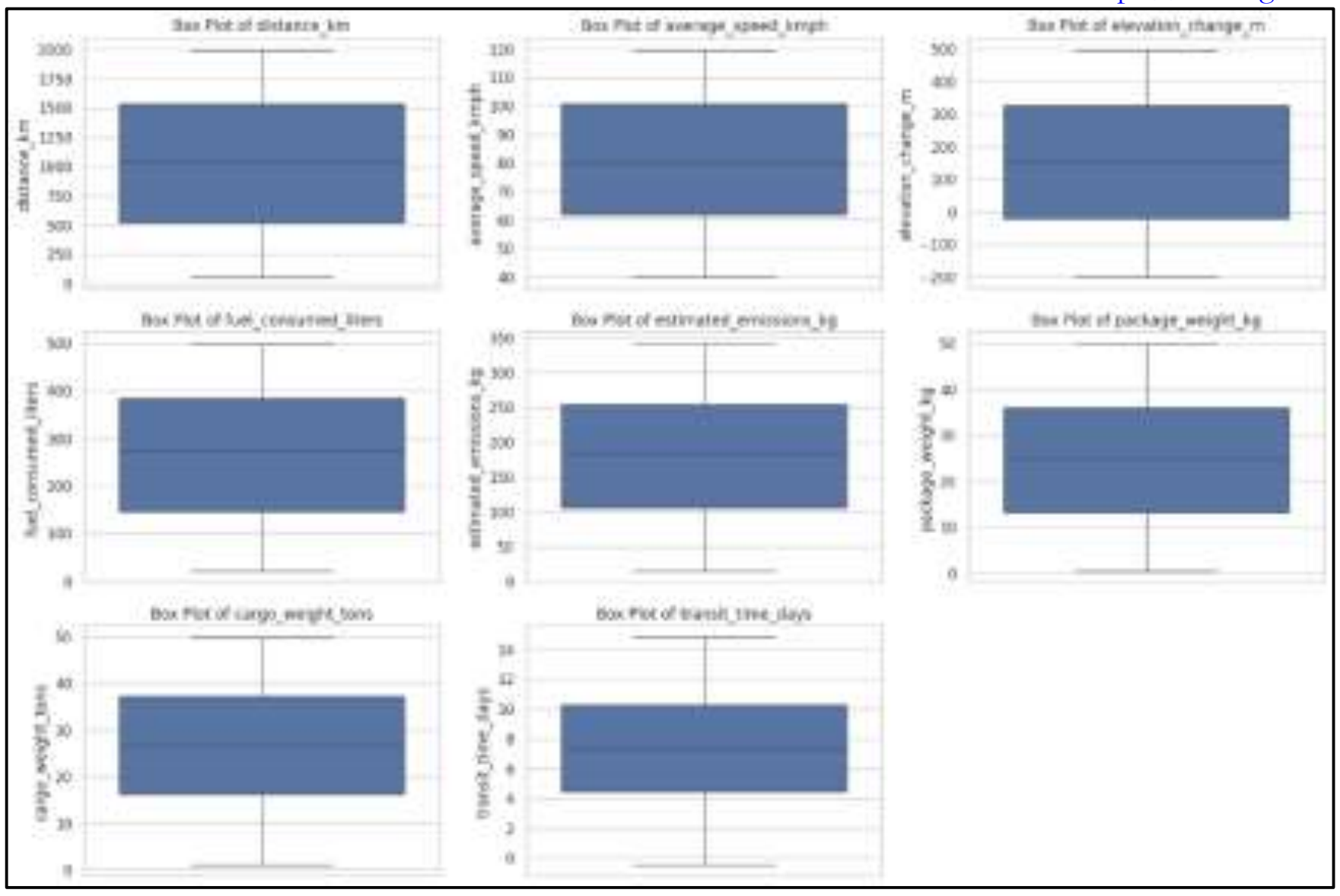

**Figure 4.** Variability and Outliers Analysis

*Average Emissions by Transport Mode*

Air transport is responsible for the highest emissions of $CO_2$ when compared to other modes of transport, with rail, ship, and truck following behind. While the differences in emissions among these modes are not extreme, air transport consistently stands out for its higher levels of emissions. The primary reason for this is that air transport burns more fuel per kilometer per unit of cargo compared to other forms of transport. In contrast, rail and shipping methods are generally more fuel-efficient, as they can transport larger volumes of goods at once. Although trucks emit relatively less $CO_2$ per unit distance traveled, they remain significant contributors to overall emissions due to their widespread use in logistics and transportation networks.

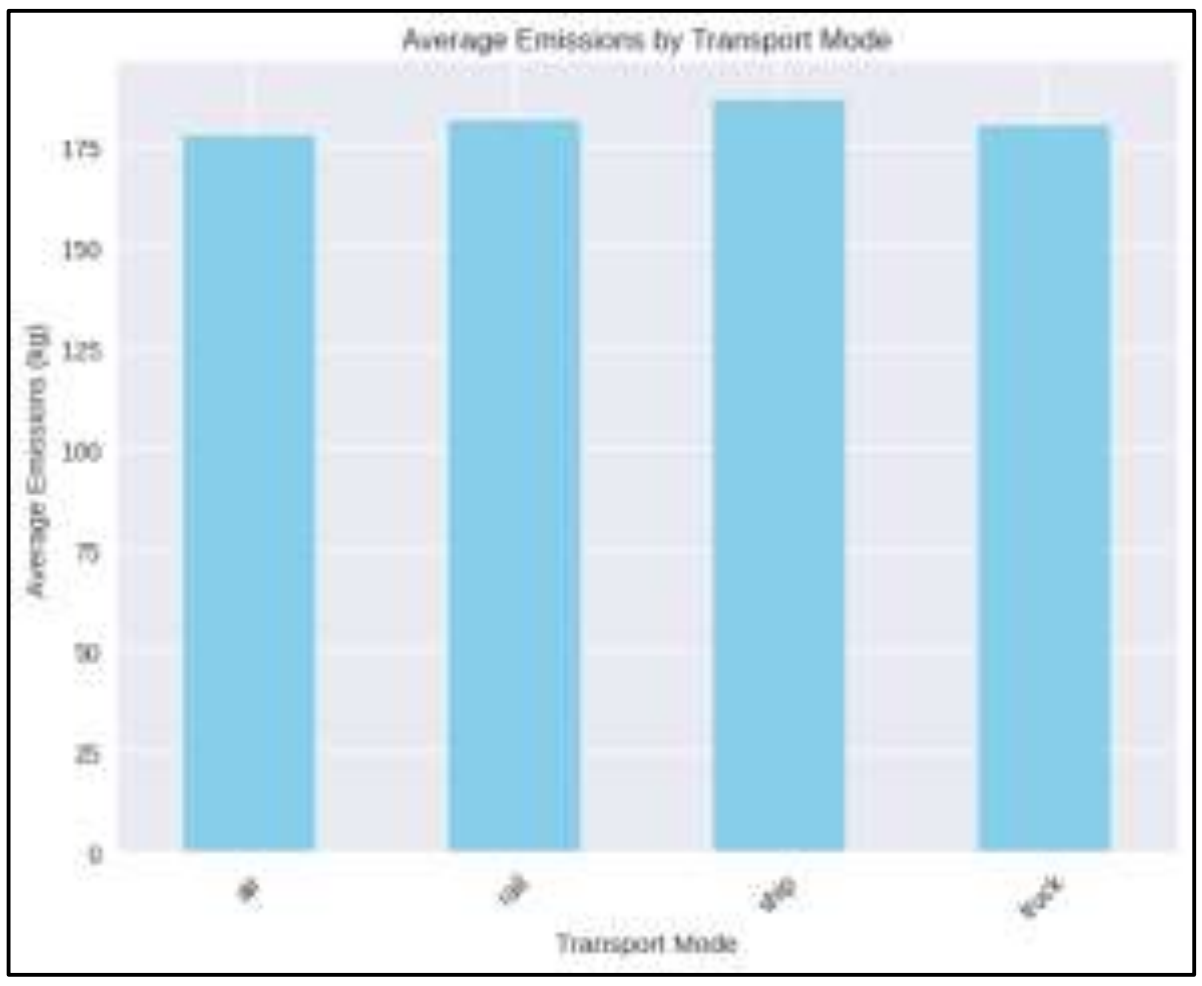

**Figure 5.** Average Emissions by Transport Mode

*Distribution of Emissions by Priority Level*

The analysis reveals that emissions are comparable across all priority levels (1, 2, and 3), as evidenced by the nearly identical interquartile ranges and medians. Notably, some shipments across all priority levels exhibit very high emissions, suggesting the presence of outliers. It is important to note that priority level does not have a direct impact on emissions. While faster deliveries, which are typically categorized as higher priority, may often utilize transport modes with higher emissions, such as air travel, this factor alone does not primarily drive emissions levels. Instead, distance and fuel consumption emerge as the key contributors to emissions, overshadowing the influence of priority.

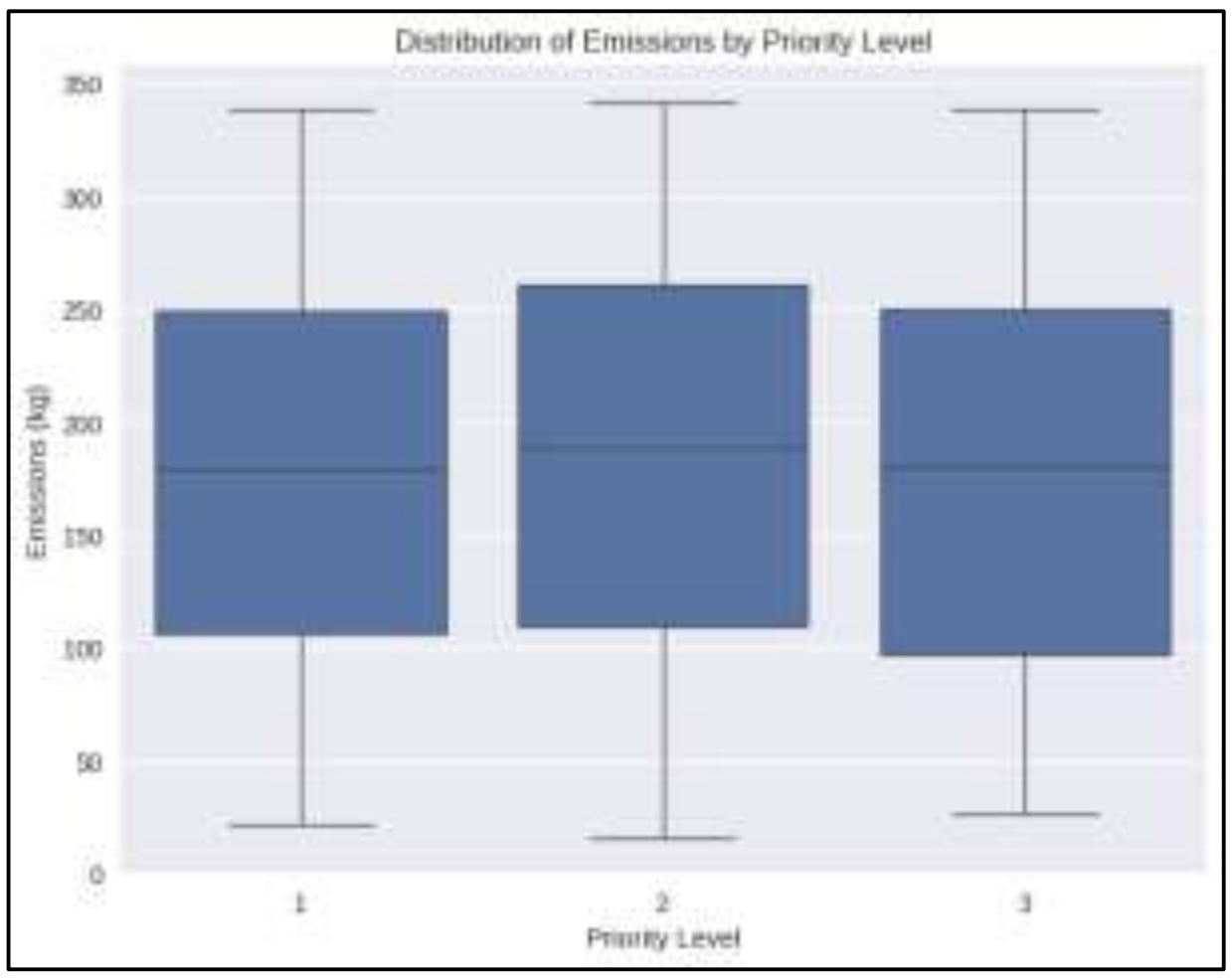

**Figure 6.** Distribution of Emissions by Priority Level

*Distance vs. Emissions by Transport Mode*

A strong linear correlation exists between distance and emissions, with all transport modes—truck, air, rail, and ship—following the same increasing trend. This relationship can be attributed to the fact that longer distances necessitate the consumption of more fuel, which in turn results in higher emissions. While it's true that different transport modes may have varying emissions per kilometer, the overall trend remains consistent and linear across all modes. Furthermore, the absence of major deviations indicates a stable and consistent level of fuel efficiency across these transport options.

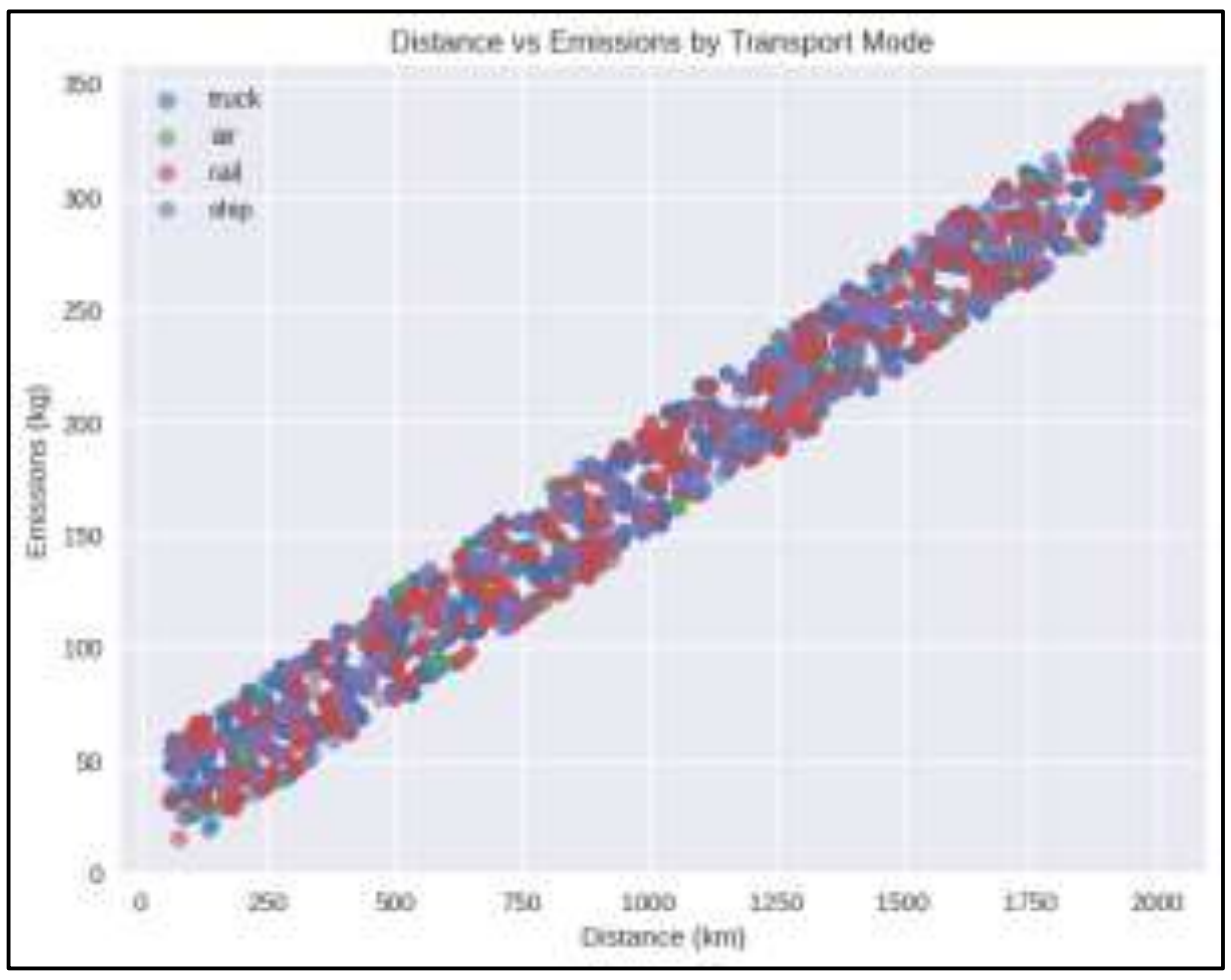

**Figure 7.** Distance vs. Emissions by Transport Mode

*Average Emissions by Fuel Type*

Aviation fuel is associated with the highest emissions among various transport fuels, primarily due to its substantial carbon footprint per unit burned. In comparison, diesel and electric transport exhibit similar emission levels. Diesel trucks and trains, while they contribute significant emissions, often benefit from better fuel efficiency, which can mitigate their overall impact to some extent. It's important to note that electric transport, despite being viewed as a cleaner alternative, is not entirely emission-free. The indirect emissions from electricity generation play a crucial role in this, as the environmental impact varies based on the energy sources used, such as coal, hydro, and others.

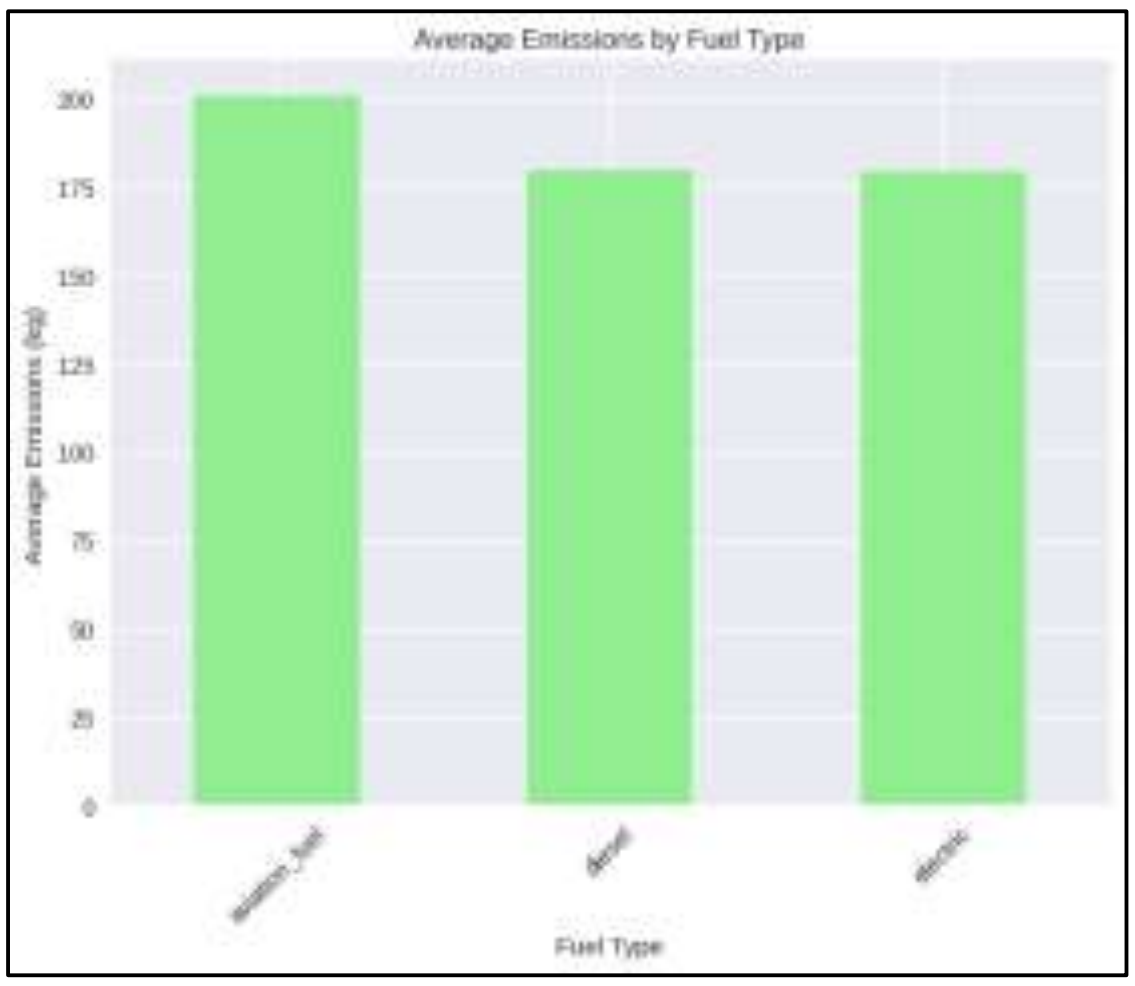

**Figure 8.** Average Emissions by Fuel Type

*Average Fuel Consumption by Vehicle Type*

Cargo ships and diesel trucks are the biggest consumers of fuel in the transportation sector. This high fuel consumption can be attributed to cargo ships operating over long distances while carrying heavy loads, which significantly increases their fuel use. Diesel trucks, commonly used for land transport, also rely heavily on fuel, contributing to their substantial overall consumption. In contrast, electric trucks, while consuming less fuel in liters, have an energy efficiency that heavily depends on the sources of electricity that power them. Freight trains, on the other hand, are known for their fuel efficiency, thanks to optimized rail transport systems that minimize energy use while maximizing cargo capacity.

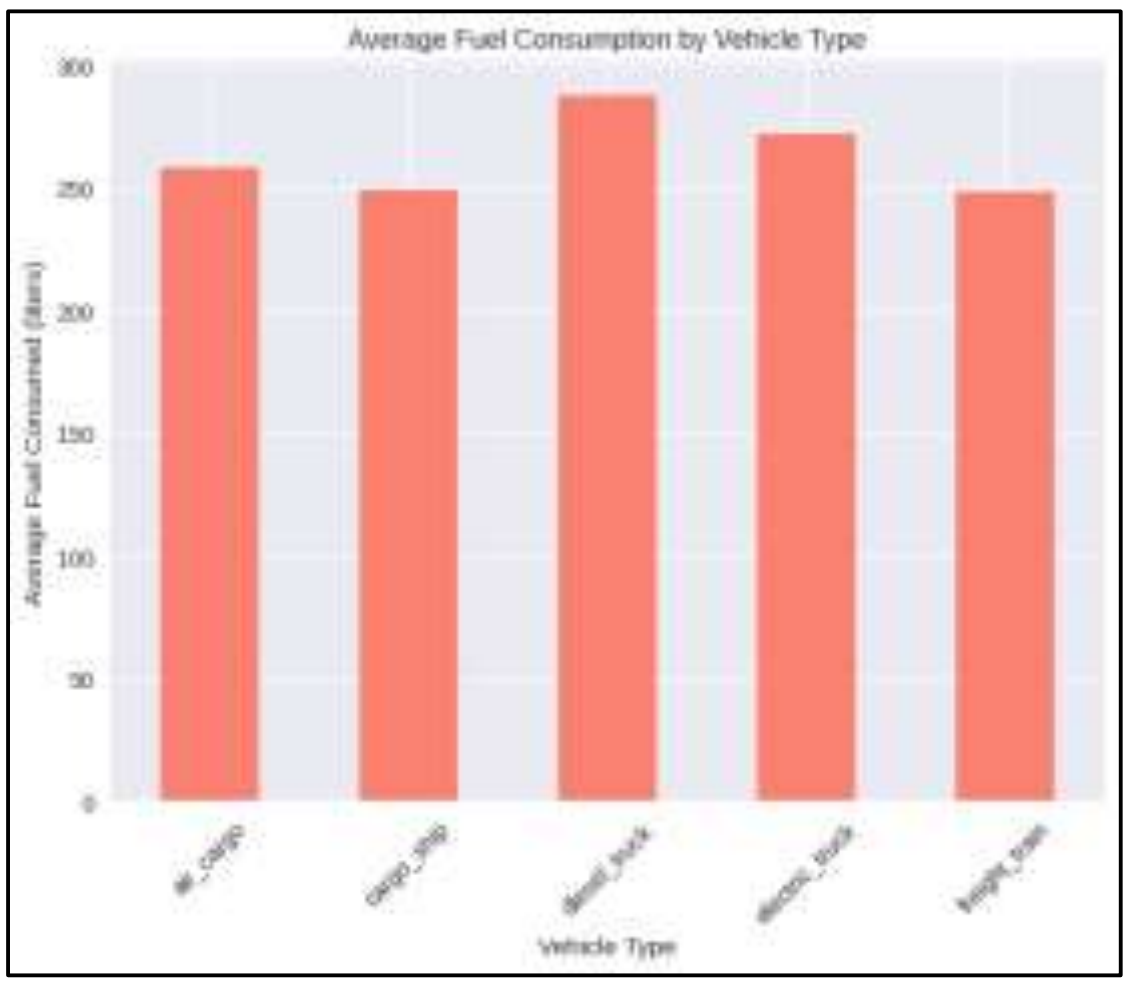

**Figure 9.** Average Fuel Consumption by Vehicle Type

*Emissions by Transport Mode and Fuel Type*

The analysis reveals several key observations about emissions across different transport modes. Trucks and ships exhibit similar emissions whether powered by electric or diesel fuel, while rail transport shows the highest emissions when utilizing aviation fuel. Interestingly, air transport powered by aviation fuel demonstrates a significant reduction in emissions compared to its electric and diesel counterparts; however, the variability in these emissions is high, as indicated by the large error bars. This variability could stem from factors such as flight distance, aircraft efficiency, and operational conditions. Notably, rail transport running on aviation fuel exceeds an average of 220 kg in emissions, indicating that aviation fuel is neither an efficient nor a clean choice for this mode of transportation, potentially due to the combustion characteristics and energy inefficiencies of locomotives designed for other fuels.

In contrast, diesel and electric transport modes maintain consistent emissions with only minor variations, suggesting similar efficiencies and fuel-to-emission conversion rates across diesel-powered trucks, rail, and ships. The presence of error bars highlights the fluctuation in emissions, particularly for air transport using aviation fuel, which varies significantly depending on factors such as flight conditions and aircraft type. This analysis leads to several implications for future policies and practices. If the aim is to reduce emissions, prioritizing electric and diesel options over aviation fuel for rail transport is advisable. Additionally, aviation fuel should primarily be reserved for air transport, where it is more effective. Encouraging investment in electric alternatives for trucking and shipping could also contribute to further emission reductions. Overall, the findings indicate that aviation fuel is the least efficient option for rail transport, while electric and diesel alternatives offer more consistent emissions performance. The substantial variability in aviation fuel emissions within air transport warrants further investigation into the influencing operational factors.

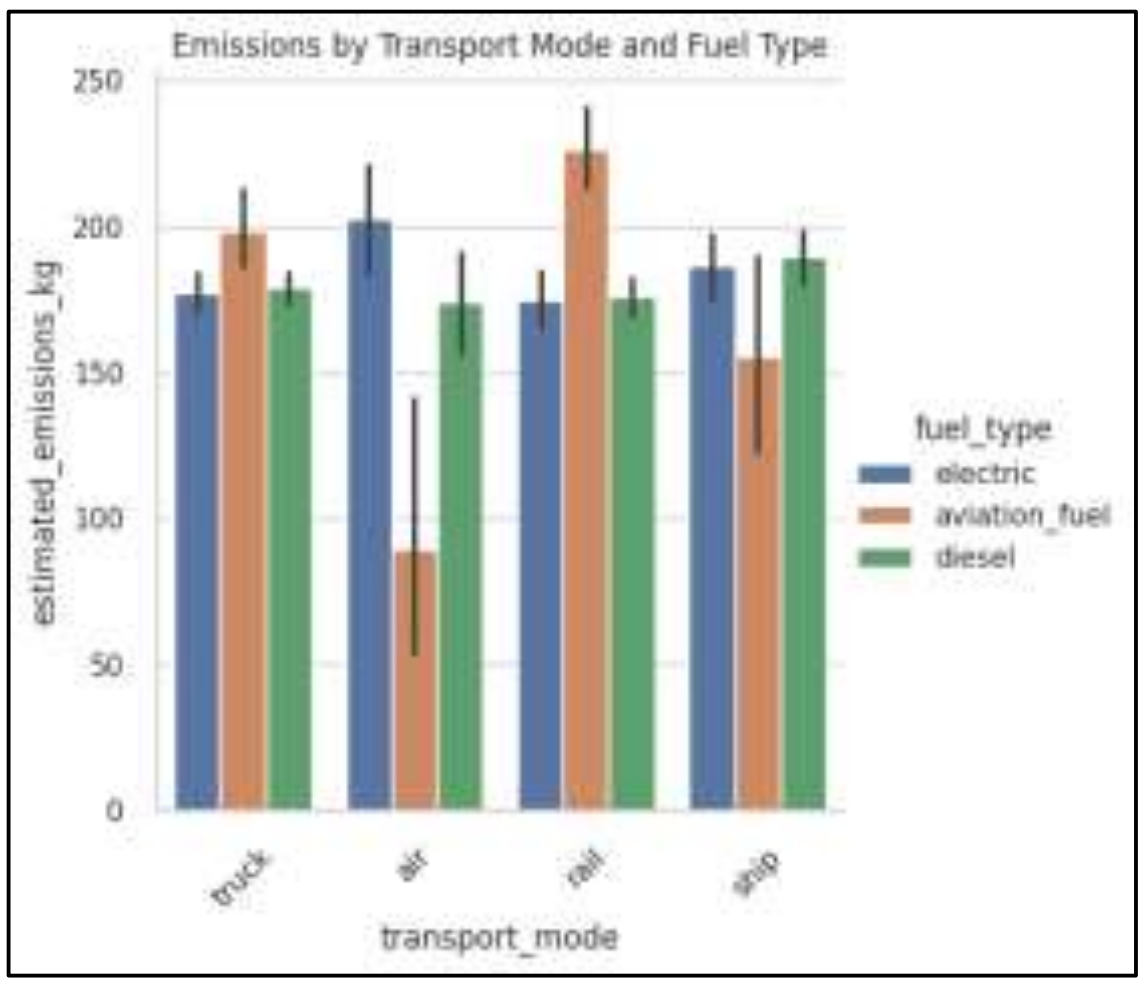

**Figure 10.** Emissions by Transport Mode and Fuel Type

*Model Development*

The research study employs a diverse array of advanced artificial intelligence and machine learning models, including Linear Regression, XGBoost, Support Vector Machines, Neural Networks, K-Means, and DBSCAN. Each of these sophisticated models is carefully crafted to tackle unique facets of logistics optimization, enhancing overall efficiency and performance in the field. Regression models, for instance, play a critical role in demand forecasting, allowing businesses to accurately predict future requirements and streamline operations accordingly. Clustering algorithms such as K-Means and DBSCAN are instrumental in optimizing delivery routes, ensuring that resources are utilized effectively to reduce operational costs. K-Means clusters the travel routes to group deliveries based on distance, traffic and travel time. On the other hand DBSCAN is used to detect outliers in Transit time to objectively detect outlier routes with inefficient delivery times. Hyperparameter tuning and model stacking was also implemented to improve the performance of the models. ***Figure 11*** displays the comparison in performance of different models after hyperparameter tuning, stacking and Feature Engineering. To determine the optimal number of clusters for the K-Means algorithm, the elbow method was utilized, which revealed that three clusters were most appropriate (as depicted in ***Figure 12***).

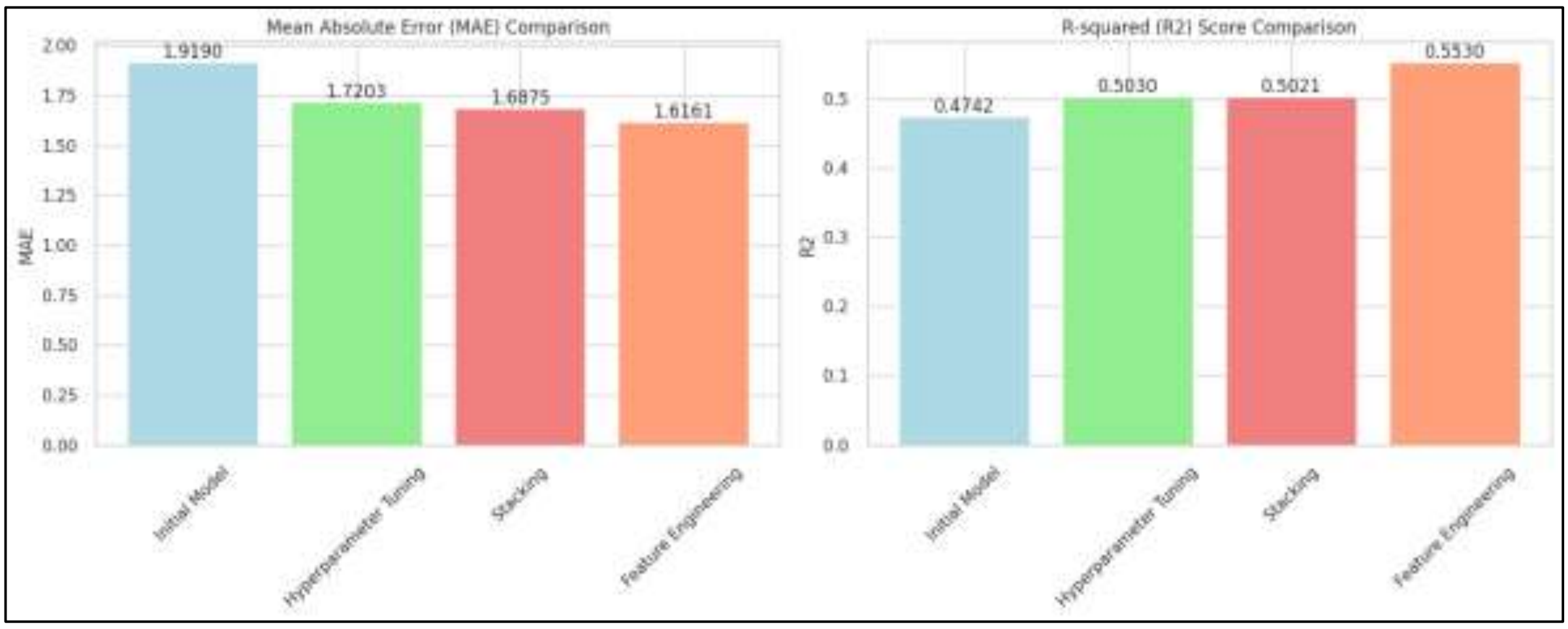

**Figure 11**. Performance of different models after hyperparameter tuning, stacking, and Feature Engineering.

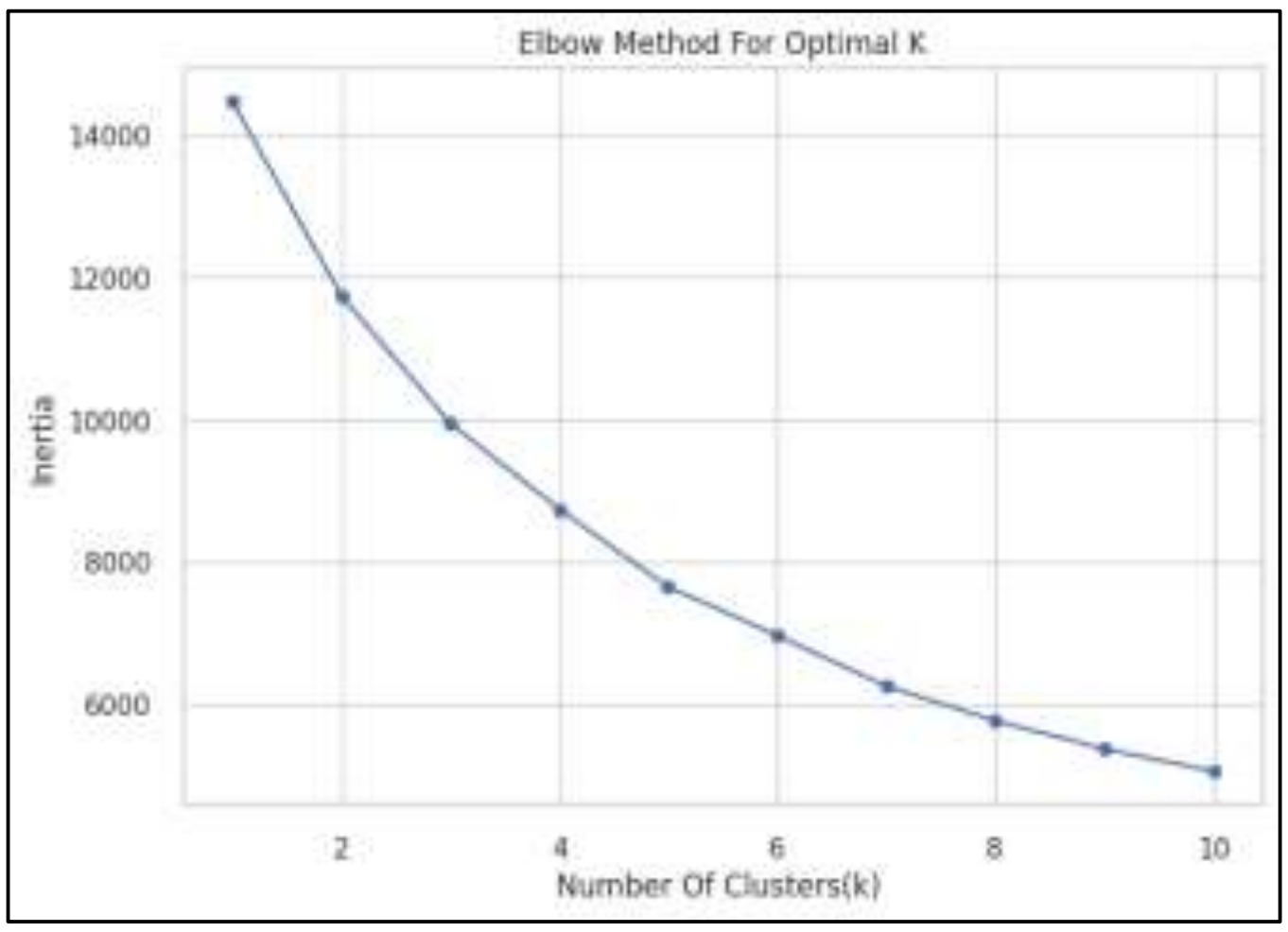

**Figure 12**. Elbow Method to determine the number of clusters.

Feature importance analysis is also carried out for the XGBoost model(***Figure 13***). The key observations from the analysis highlight the most important features influencing the target variable, particularly emphasizing that fuel consumption (in liters) is the most significant predictor. This indicates a direct correlation between fuel use and the outcome, likely relating to emissions or transport costs. Vehicle type, specifically air cargo, also plays a crucial role in the model's predictions, likely due to its high operational costs and fuel consumption. Additionally, traffic impact scores and elevation effects were found to significantly influence the model, as they are likely to affect travel time, fuel usage, and emissions.

Moderately important features include elevation change, which is vital in determining fuel consumption and transport efficiency, as well as fuel type—whether diesel, electric, or aviation fuel—which significantly impacts transport efficiency and emissions, more so than certain transport modes. Cargo weight is another important factor, as heavier loads increase transport emissions and energy usage. Traffic level and average speed also contribute to route efficiency and fuel consumption. On the other hand, features like priority and package weight have been identified as the least important, indicating that specific package details have minimal influence when compared to general cargo weight and fuel consumption. The impacts of diesel and electric trucks are relatively lower than those of other transport types, such as air cargo and ships.

The implications are clear: strategies should focus on optimizing routes to reduce fuel consumption, which is crucial for enhancing transport efficiency and lowering emissions. Planning routes must account for elevation changes and traffic conditions to maximize fuel efficiency. Furthermore, considering cleaner fuel types, like electric or hybrid options, could greatly enhance sustainability efforts. In conclusion, the analysis shows that fuel consumption, the mode of transport (especially air cargo), and the impacts of traffic and elevation are the primary predictors in the model. This underscores the importance of efficient fuel management, optimized routing, and thoughtful transportation mode choices in improving logistics and minimizing environmental impact.

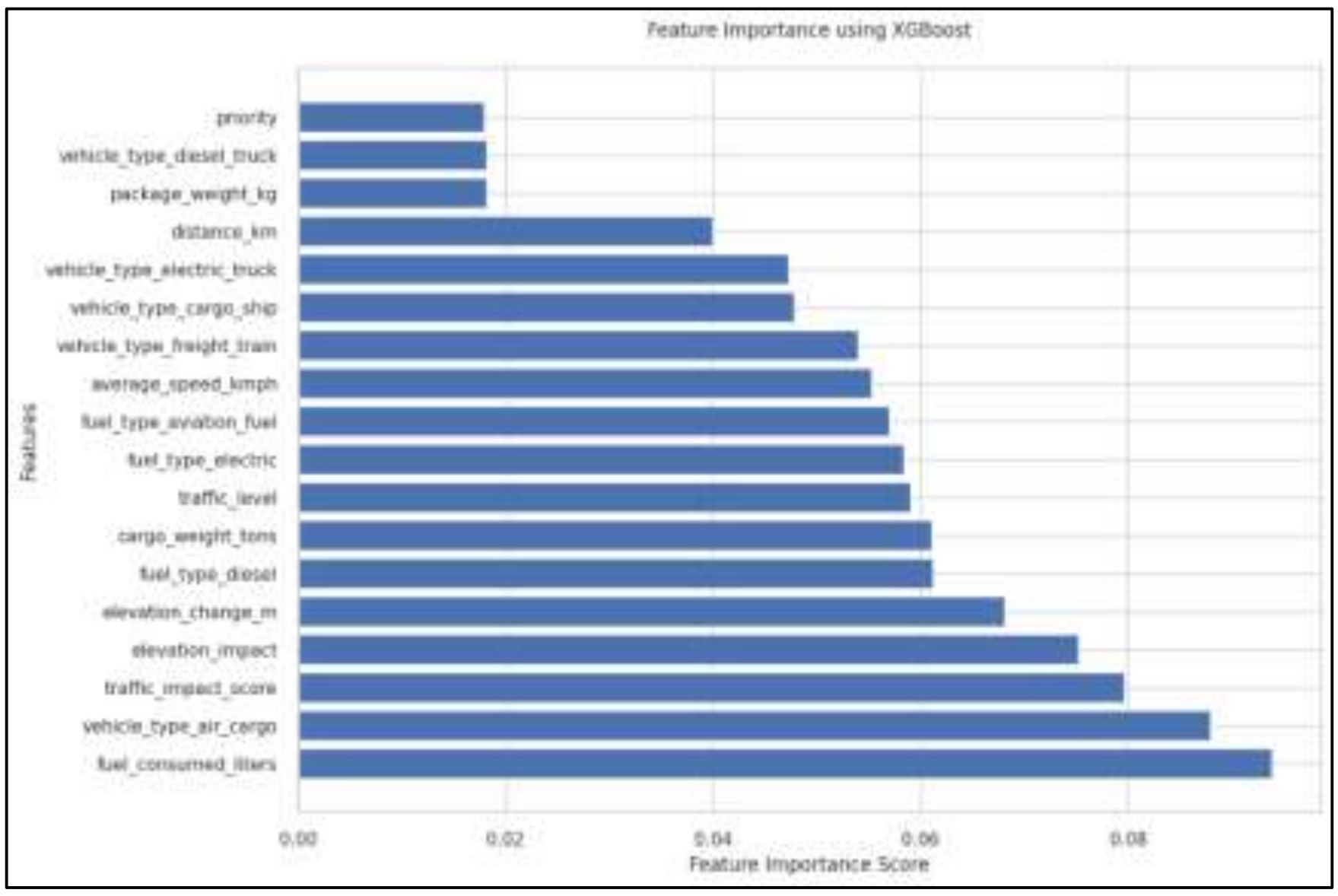

**Figure 13.** Feature importance analysis of the XGBoost Model

*Model Training and Validation Procedures*

The process of model training unfolds in a meticulously structured manner, commencing with the careful partitioning of data into three distinct sets: training, validation, and testing. The training set serves as the foundation for developing the model, providing the essential data needed to fit and optimize its parameters. In contrast, the validation set plays a pivotal role in the fine-tuning process, allowing for hyperparameter adjustments and performance evaluations that enhance the model's predictive capabilities. To ensure the robustness and reliability of the model, cross-validation techniques, such as k-fold cross-validation, are employed. This method divides the training data into multiple subsets, enabling the model to be tested on different combinations of data and further solidifying its generalizability. Once the models have been refined, their final performance is meticulously assessed using the test set—a separate dataset reserved solely for this purpose. Key metrics such as mean absolute error (MAE), mean squared error (MSE), and the $R^2$ score are utilized to quantify the model's accuracy and effectiveness in making predictions. Throughout this iterative process, models are continuously honed and enhanced, ensuring they not only achieve high accuracy but also remain adaptable and reliable when deployed in real-world scenarios.

## Results and Evaluation

*High Carbon Emissions*

The main objective for model development was to predict and reduce carbon emissions based on logistics operations. From the dataset, it is observed that Rail transport utilizing aviation fuel seems to be inefficient regarding emissions. In contrast, air transport significantly benefits from aviation fuel, as it plays a crucial role in reducing emissions during flight. Electric transport is consistently making strides in emission reductions; however, its effectiveness can differ across various transport modes. Additionally, ships and trucks exhibit comparable emission levels, indicating a potential opportunity for optimization in these areas to further enhance sustainability(***Figure 14***).

The primary models trained for mitigating carbon emissions include Linear Regression, Multilayer Perceptron Regressor (MLPRegressor), XGBoost Regressor (XGBRegressor), and Random Forest Regressor. Among these four models, XGBRegressor emerged as the top performer, achieving a Mean Absolute Error of 0.454 and an impressive R-squared score of 0.999, closely followed by RandomForestRegressor.

**Table 1**. Model Performance comparison for Carbon Emissions.

| Model | MAE | R-Squared |
|---|---|---|
| Linear Regression | 11.026 | 0.977 |
| MLP Regressor | 10.662 | 0.979 |
| Random Forest Regressor | 0.579 | 0.999 |
| XGB Regressor | 0.454 | 0.999 |

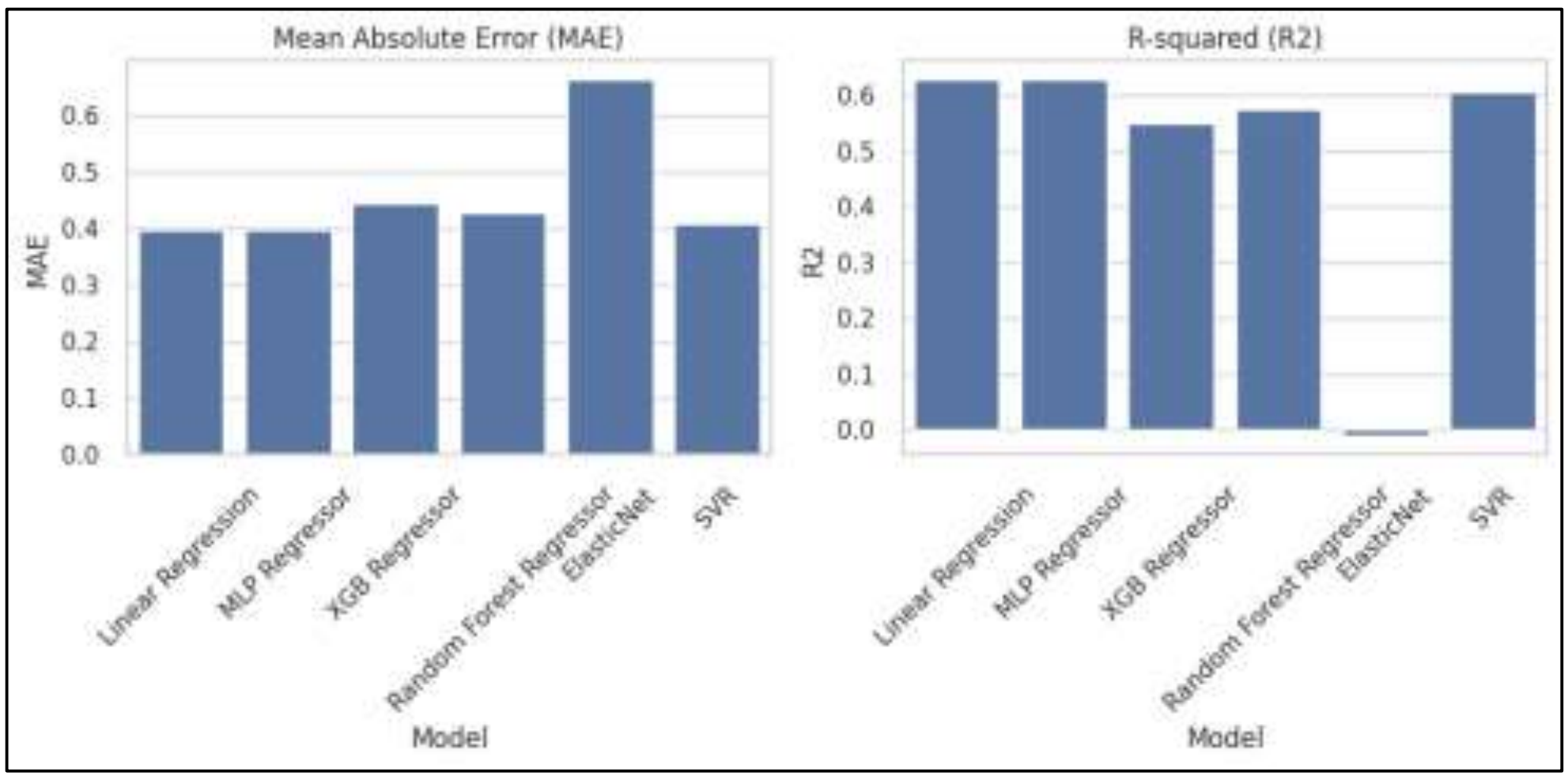

**Figure 14.** Performance comparison for carbon emissions prediction models.

*Optimizing Travel Routes*

Five machine learning models were trained with the primary goal of optimizing travel routes to minimize distance and travel time. The models used in this study include Linear Regression, MLPRegressor, XGBRegressor, ElasticNet, and SVR. Among these, XGBRegressor emerged as the best-performing model, achieving a Mean Absolute Error of 1.919 and an R-squared score of 0.474. RandomForestRegressor followed closely, while Linear Regression was the poorest-performing model in this analysis(***Figure 15***).

**Table 2.** Model Performance Comparison of Travel Route Optimization.

| Model | MAE | R-Squared |
|---|---|---|
| Linear Regression | 3.125 | -0.002 |
| MLP Regressor | 3.157 | -0.018 |
| Random Forest Regressor | 2.016 | 0.471 |
| XGB Regressor | 1.919 | 0.474 |
| Elastic Net | 3.087 | -0.001 |
| SVR | 3.048 | 0.023 |

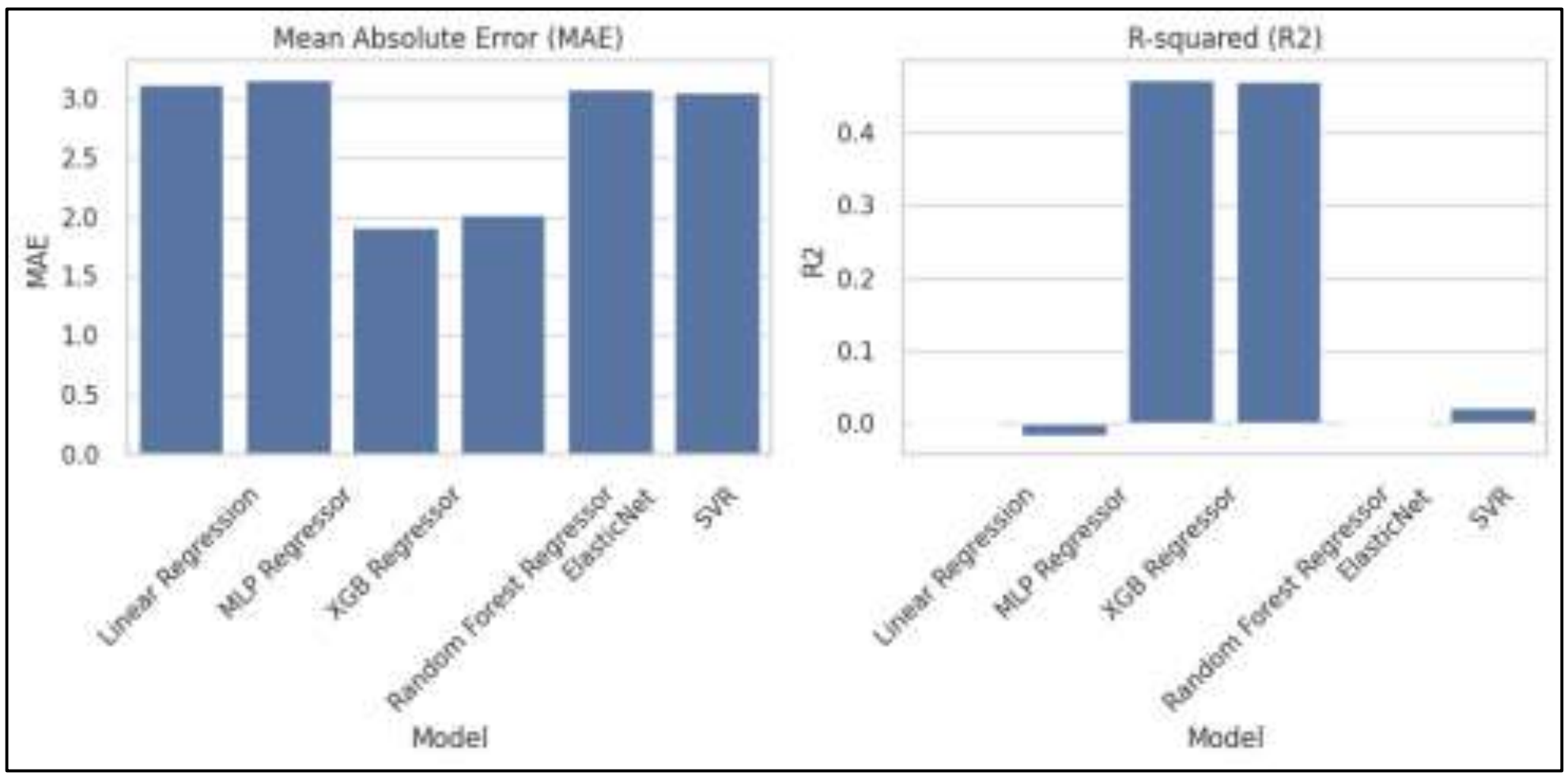

**Figure 15.** Performance comparison of Travel Route Optimization models.

*Route Clustering*

The K-Means algorithm model was employed to effectively group deliveries based on key variables such as distance, traffic conditions, and travel time. After fitting the K-Means algorithm to the dataset, the distinct clusters emerged as illustrated in ***Figure 16***. The first cluster, designated as cluster 0, is characterized by high traffic levels coupled with relatively shorter distances. This combination suggests that routes within this cluster are plagued by dense urban traffic, which significantly delays deliveries despite their shorter length. Such conditions are common in metropolitan areas, where constant congestion can impede even the most direct routes. In contrast, the second cluster, known as cluster 1, showcases a different scenario—high speed paired with longer distances. This indicates that, although the routes are longer, they permit faster travel likely due to fewer interruptions and minimal traffic congestion. These routes are typically found in rural areas where open roads allow for smoother transit without the hindrance of urban traffic jams. The final cluster, cluster 2, presents the challenges of longer travel distances combined with slower speeds. This cluster indicates routes that experience the longest transit times, possibly due to difficult terrains, such as mountainous regions, or areas with inadequate infrastructure. Deliveries in this cluster are often hampered by logistical challenges that necessitate careful planning.

For the first cluster, it is advisable to prioritize the use of smaller, agile vehicles that can adeptly maneuver through heavy traffic. Scheduling deliveries during off-peak hours can significantly alleviate congestion, and using real-time traffic applications for rerouting can further optimize travel times. In the case of the second cluster, deploying larger trucks or freight trains for bulk deliveries would be effective. Additionally, optimizing fuel efficiency by utilizing diesel-powered vehicles can markedly enhance operational efficacy, ensuring that deliveries remain on schedule despite the longer distances involved. Finally, for the last cluster, it's crucial to allocate extra time buffers to account for potential delays. Using robust vehicles that are designed to navigate rough terrains will be essential in these challenging environments. By implementing these targeted strategies, we can enhance resource allocation and improve overall delivery efficiency across diverse routing scenarios.

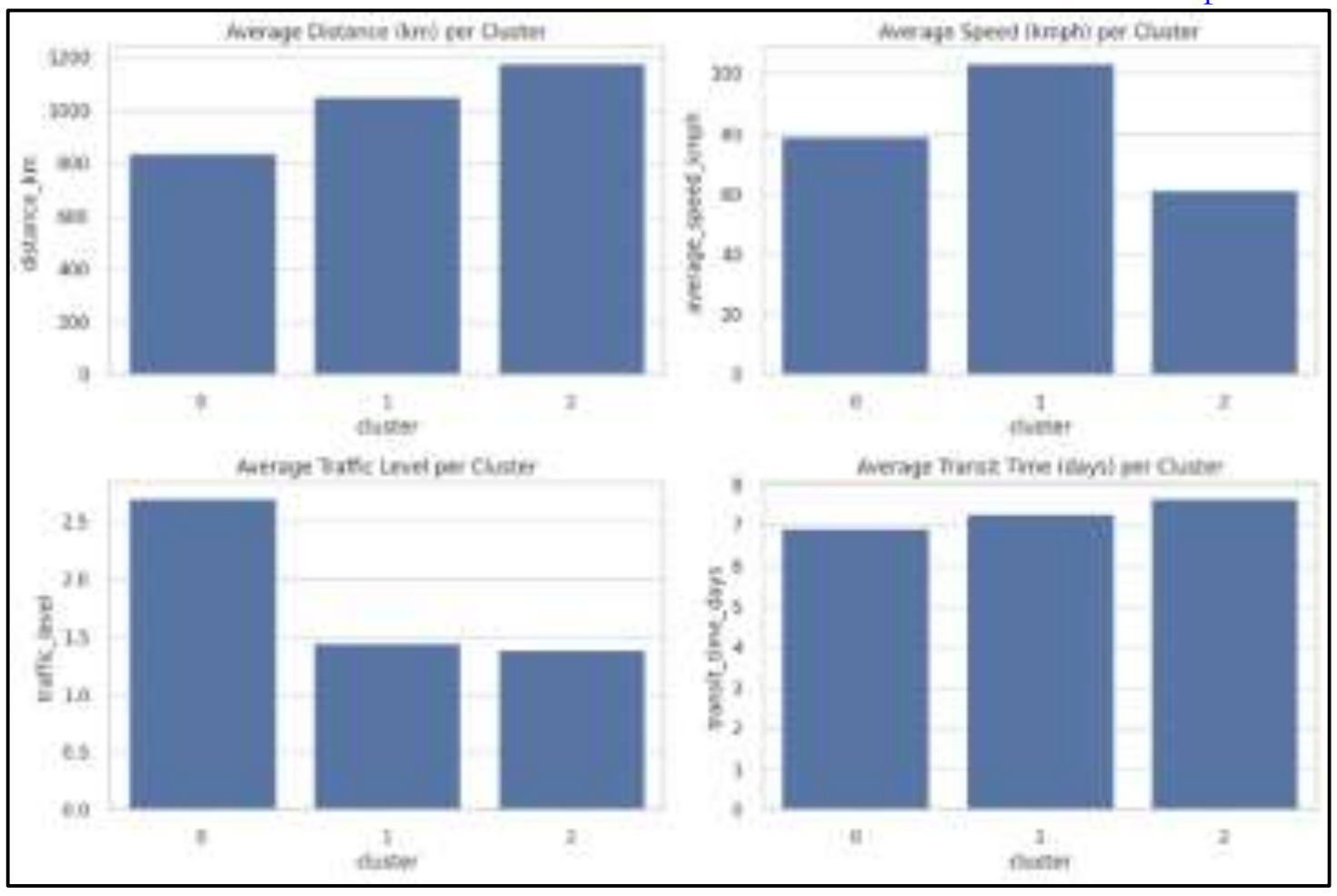

**Figure 16.** Cluster-wise Analysis of Transportation Metrics: Distance, Speed, Traffic Level, and Transit Time

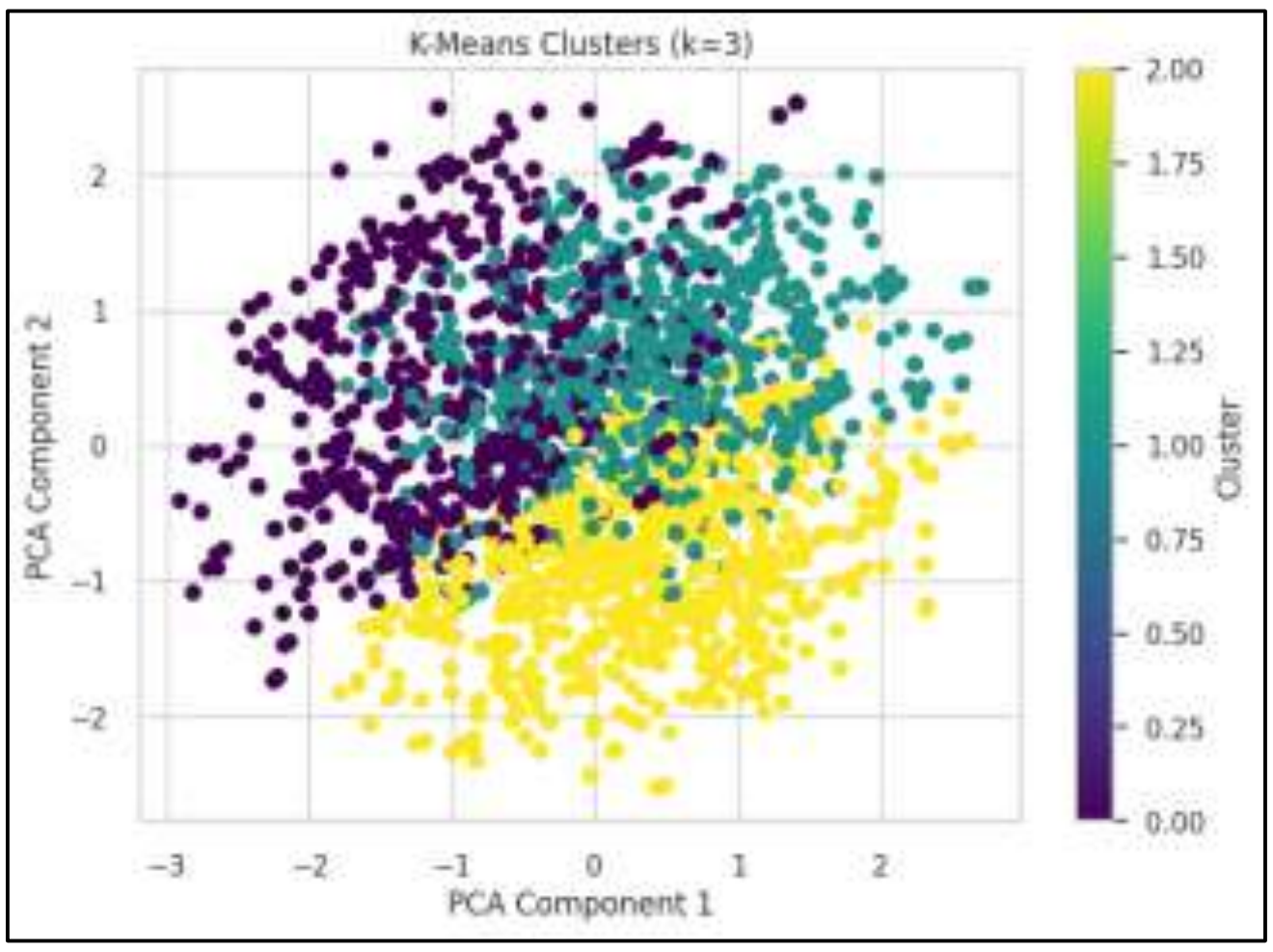

**Figure 17.** Visualization of the 3 clusters using PCA after applying K-Means clustering

*Outlier Detection*

The DBSCAN model is utilized to detect outliers in travel time. To determine the epsilon value for the DBSCAN model we compute the k-Distance Graph graph(NearestNeighbors). The epsilon value represents the maximum distance between two points for them to be considered neighbours. A total of 48 routes were found to have outliers based on unusual distances, abnormal transit times, and extreme combinations of distance and transit times. Outlier investigation can be carried out and abnormal values be removed. These abnormal values could be clearly incorrect values and could be removed using Interquartile Range(IQR Method) or Z-Score. In the scatter plot(***Figure 18***), the data points are color-coded to indicate their status as outliers or inliers. Red dots highlight the strongest outliers, signifying the most extreme anomalies in the dataset. In contrast, light orange or beige dots represent less extreme anomalies, showing a more moderate deviation from the norm. Meanwhile, dark blue dots identify the inlier data points, which are considered normal within the context of the data.

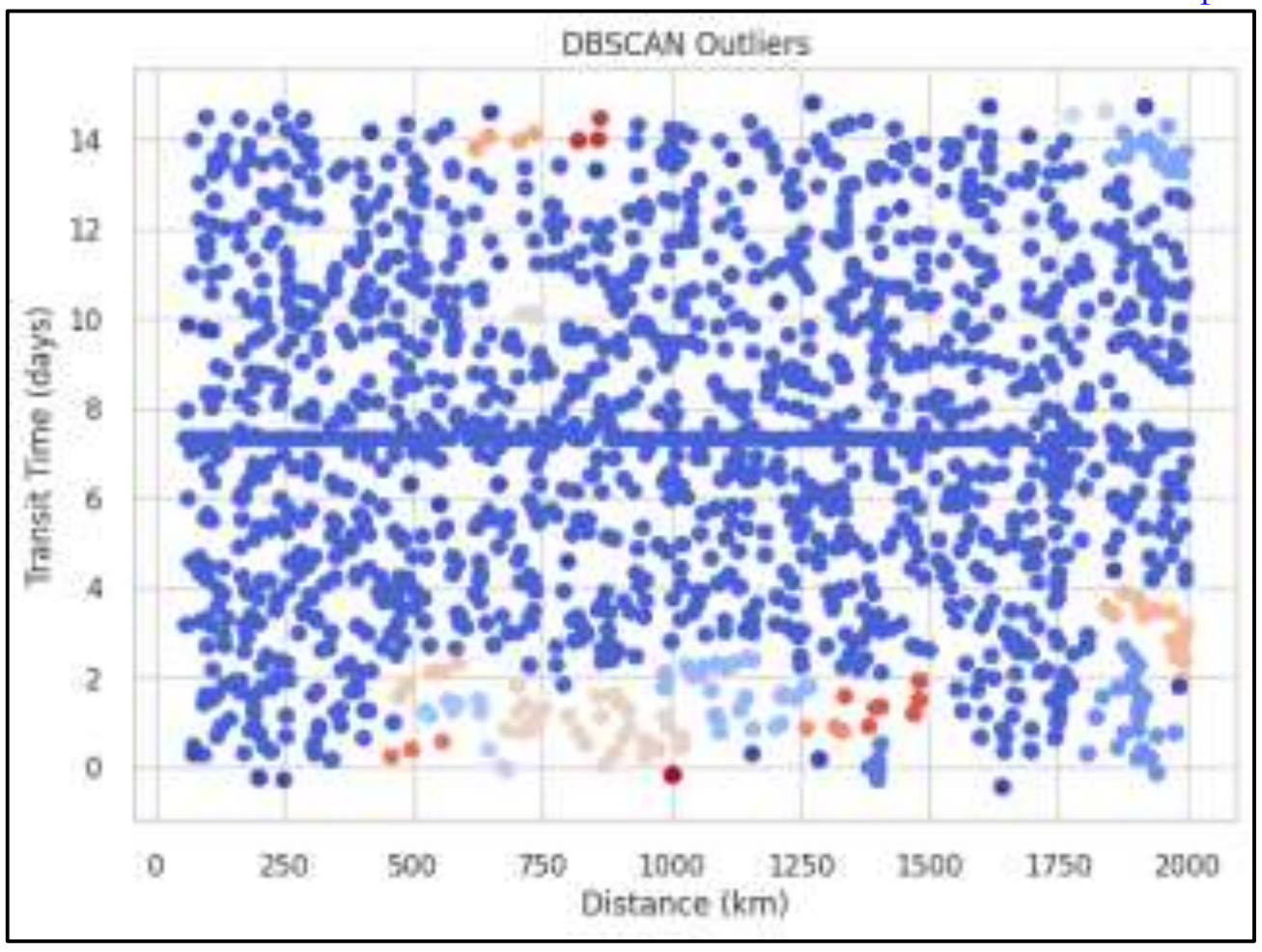

**Figure 18**. This color scheme effectively distinguishes between various levels of deviation in the outliers

*Demand Forecasting*

In the case of demand forecasting, Linear Regression, MLPRegressor, XGBRegressor, RandomForestRegressor, ElasticNet, and SVR models are utilized to prognosticate forthcoming deliveries to strategize optimal routes. The MLPRegressor and Linear Regression are the best-performing models with ElasticNet being the poorest performer. The MLPRegressor, which is a neural network featuring three hidden layers with eleven neurons each, shows a slight improvement over Linear Regression. This advantage comes from the non-linearity inherent in neural networks, enabling them to capture more complex relationships within the data. As a result, the $R^2$ value increases from 0.6295 to 0.6349, indicating that deep learning provides a marginal benefit in terms of model performance.

**Table 3**. Model performance comparison of demand forecasting models

| Model | MAE | R-Squared |
|---|---|---|
| Linear Regression | 0.398 | 0.629 |
| MLPRegressor | 0.396 | 0.635 |
| RandomForestRegressor | 0.434 | 0.566 |
| XGBRegressor | 0.450 | 0.542 |
| ElasticNet | 0.664 | -0.011 |
| SVR | 0.423 | 0.573 |

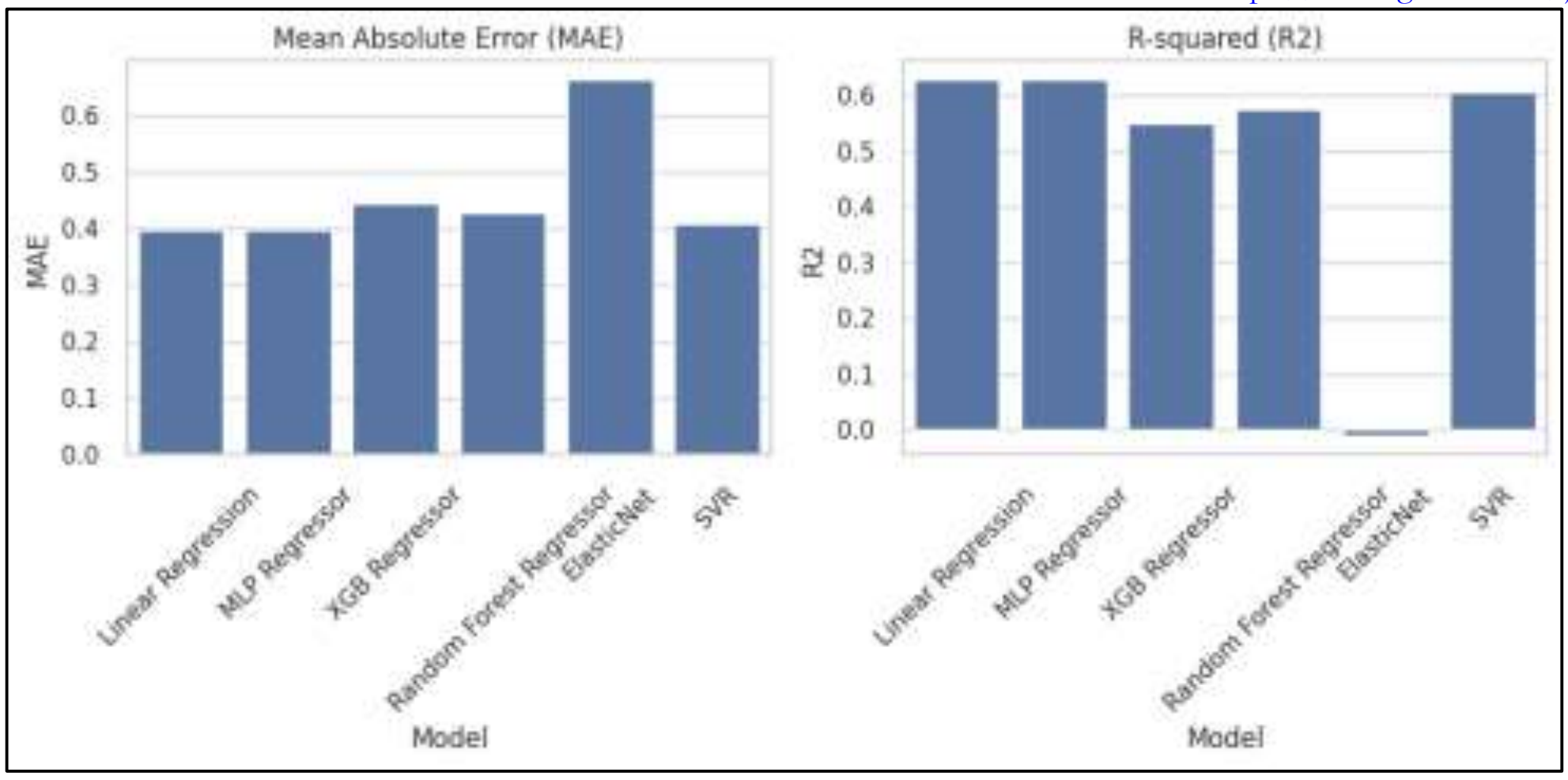

**Figure 19**. Performance comparison of demand prediction models.

*Deployment and Real-World Applications*

*Model Deployment Strategies*

Cloud platforms such as Amazon Web Services (AWS), Google Cloud Platform (GCP), and Microsoft Azure offer robust and scalable solutions tailored for deploying AI models in the logistics sector. One of the standout features of cloud deployment is elastic scalability, which empowers logistics companies to effortlessly manage varying demand levels by adjusting their computational resources in real time. This capability ensures that organizations can efficiently allocate resources during peak times without incurring unnecessary costs during slower periods. Additionally, the remote accessibility provided by these platforms allows logistics professionals to access critical real-time analytics from any location, enhancing decision-making and enabling more responsive operations. Moreover, the seamless integration of cloud services with Internet of Things (IoT) devices and edge computing technologies facilitates precise real-time tracking of fleet movements and inventory levels. This interconnectedness not only optimizes supply chain visibility but also supports proactive maintenance and operational adjustments, leading to improved efficiency and customer satisfaction throughout the logistics process.

Edge computing plays a crucial role in real-time logistics applications, particularly where latency is a significant concern. Rather than depending on cloud servers, AI models are deployed directly on edge devices, such as GPS-enabled logistics trackers, IoT sensors, and autonomous vehicles. This approach offers several benefits, including low latency processing, which significantly reduces response times in dynamic logistics environments. Additionally, edge computing ensures offline functionality, allowing models to operate effectively even in areas with limited internet connectivity. Moreover, it enhances security by processing sensitive data locally, thereby reducing the risk of data breaches. Overall, the integration of edge computing in logistics improves efficiency and resilience in managing complex operations.

Hybrid deployment effectively combines cloud and edge computing to achieve a balance between scalability and real-time processing. In this setup, AI models are utilized to make critical decisions at the edge, while the cloud is leveraged for advanced analytics and model updates. One of the prominent use cases of this approach is predictive maintenance, where AI models deployed on trucks analyze real-time sensor data and sync with cloud-based predictive maintenance systems. Additionally, dynamic route optimization benefits from this hybrid model, as it merges real-time edge-based decision-making with historical data analysis sourced from the cloud. This integration enhances operational efficiency and responsiveness in various applications.

For logistics platforms that aim to integrate multiple AI solutions, adopting an API-based deployment strategy proves to be highly effective. By deploying AI models as RESTful APIs, third-party logistics management systems (LMS) can seamlessly access predictive analytics. This approach offers several key advantages, including interoperability, which facilitates integration with various logistics software platforms. Additionally, it simplifies maintenance since model updates can be implemented without disrupting the core logistics infrastructure. Moreover, the integration of APIs into existing transport management systems (TMS) allows for automated decision-making, enhancing processes such as routing and scheduling.

*Real-World Applications*

UPS(United Parcel Service), a global logistics and package delivery company based in the United States, employs an AI-powered logistics optimization system called ORION (On-Road Integrated Optimization and Navigation), which utilizes machine learning to enhance its delivery processes. This advanced system minimizes travel distances, leading to a reduction in fuel consumption and emissions. Additionally, ORION dynamically reroutes deliveries based on real-time traffic conditions, ensuring that packages reach their destinations efficiently. As a result, UPS saves millions of gallons of fuel each year, which significantly contributes to lowering their overall carbon footprint.

Amazon has integrated AI-driven logistics solutions to enhance its warehouse management and last-mile delivery operations. A key application of this technology is Robotic Process Automation (RPA), where AI-powered robots are utilized to optimize inventory placement and streamline order fulfillment. Additionally, the company employs dynamic pricing and demand forecasting through machine learning models, which predict fluctuations in demand, thereby improving the resilience of its supply chain. Furthermore, Amazon has advanced its delivery capabilities with the use of drones and autonomous vehicles, which are designed to reduce inefficiencies in last-mile delivery, ultimately resulting in faster and more reliable service for customers.

DHL(Dalsey, Hillblom, and Lynn), a global logistics and courier company, is integrating AI to enhance supply chain sustainability and improve operational efficiency. One of the key AI-driven solutions is predictive analytics for demand forecasting, which helps reduce overstocking and understocking of goods. Additionally, the company has implemented green logistics initiatives, where AI models optimize delivery routes to minimize carbon emissions. Furthermore, IoT-enabled fleet management utilizes AI to monitor vehicle health, thereby reducing downtime and maintenance costs. This comprehensive approach not only contributes to a more sustainable supply chain but also improves overall operational performance.

Tesla's autonomous trucking division harnesses the power of cutting-edge AI-driven Full Self-Driving (FSD) technology to revolutionize freight transportation. By minimizing human error, these intelligent, self-driving trucks not only elevate safety standards but also enhance overall operational efficiency, paving the way for a more reliable logistics landscape. The innovative AI models meticulously coordinate acceleration and braking, leading to optimized energy consumption; this process not only conserves valuable battery power but also contributes to a more sustainable transport solution. Moreover, Tesla's sophisticated AI-based fleet management system acts as the brain of the operation, ensuring smart and seamless coordination across the fleet. This strategic approach to logistics operations enables timely deliveries and maximizes the effectiveness of the entire transportation network, ultimately transforming how goods are moved across the country.

*Future Work*

As artificial intelligence (AI) continues to shape the logistics and supply chain industry, several areas require further exploration to enhance efficiency, sustainability, and adaptability. Future research should prioritize improving model accuracy, integrating AI with emerging technologies, addressing ethical and regulatory concerns, and promoting sustainable logistics practices. One key area for improvement is enhancing the accuracy and adaptability of AI models. Although AI has significantly improved logistics operations, current models still face challenges in dynamic and unpredictable environments. Future efforts should focus on improving data quality and diversity to ensure that AI systems can generalize effectively across various logistics networks. Additionally, developing self-learning AI models through reinforcement learning and self-adaptive neural networks could enable more responsive and real-time logistics optimization (Zhang et al., 2024) [24]. Reducing bias in AI models is another critical aspect, as biased data can lead to inefficiencies in delivery planning and cost predictions, ultimately impacting the overall performance of logistics systems (Miller et al., 2022) [15].

The integration of AI with emerging technologies presents another promising avenue for future research. For example, quantum computing has the potential to solve complex optimization problems in logistics much faster than traditional computing methods, which could revolutionize supply chain management (Brown et al.,2024) [3]. When integrated with AI, blockchain technology can enhance transparency and security in logistics by preventing fraud and enabling real-time tracking of shipments (Gupta et al.,2023) [5]. Moreover, with the expansion of 5G networks, AI models deployed on edge devices could process logistics data in real time, reducing latency in fleet management and warehouse automation (Williams et al., 2023) [23].

Despite these technological advancements, the adoption of AI in logistics raises several ethical and regulatory concerns. Data privacy and security are critical issues, especially as AI systems process vast amounts of sensitive logistics data. Future research should focus on ensuring compliance with data protection regulations such as GDPR and CCPA to safeguard user information (Johnson et al., 2023) [11]. Additionally, enhancing AI explainability is crucial, as logistics managers need to understand and trust AI-driven recommendations to make informed decisions (Rahman et al., 2024) [17]. Governments and policymakers must also establish clear regulatory frameworks for AI-driven logistics, particularly for autonomous vehicles and drone-based delivery systems (Nguyen et al., 2023) [16].

Sustainability is another vital consideration for future research, as AI-driven logistics solutions must align with global environmental goals. The development of energy-efficient AI models could help reduce the computational power required for logistics optimization, thereby lowering carbon footprints (Rodriguez & Kim, 2023) [18]. AI can also support circular economy logistics by optimizing product recycling and waste reduction strategies, promoting more sustainable supply chain operations (Martinez et al., 2024) [13]. Furthermore, AI-powered route planning and fleet management systems can minimize fuel consumption and emissions, contributing to environmentally friendly logistics solutions (Singh et al., 2023) [19]. Moving forward, researchers and industry stakeholders should prioritize developing standardized AI benchmarks for logistics optimization, explore AI-human collaboration models to enhance decision-making and investigate the long-term economic implications of AI-driven logistics automation.

## Conclusion

This study effectively demonstrates the significant impact of artificial intelligence (AI) and machine learning (ML) on optimizing logistics operations, reducing carbon emissions, and enhancing sustainability in the U.S. supply chain sector. By harnessing the power of predictive analytics, optimizing routing models, and utilizing AI-driven demand forecasting, businesses can achieve greater efficiency, minimize operational costs, and align their logistics strategies with environmental sustainability goals. The research utilized various AI techniques, including Linear Regression, XGBoost, Support Vector Machines, Neural Networks, and clustering algorithms such as K-Means and DBSCAN, to analyze logistics datasets. The results indicate that AI-driven models significantly improve transportation efficiency by reducing fuel consumption, optimizing delivery routes, and identifying outlier transit patterns. Additionally, applications of AI like reinforcement learning and deep learning have shown effectiveness in real-time traffic prediction and adaptive decision-making within logistics.

While this study presents compelling evidence of AI's benefits for logistics optimization, it also acknowledges several challenges that must be addressed for wider industry adoption. Key challenges include inconsistencies in data quality, high computational costs, and the integration of AI models within legacy supply chain systems. Furthermore, uncertainties surrounding regulations for AI-driven logistics solutions and the ethical implications of AI automation are areas of concern. Future research should focus on refining AI methodologies through hybrid models, integrating quantum computing for advanced problem-solving in logistics, and developing standardized benchmarks to measure AI's environmental impact. Moreover, adopting AI-powered blockchain solutions and federated learning could enhance data security and enable decentralized decision-making in logistics. Addressing these challenges is crucial for scaling AI applications and ensuring their long-term viability in global supply chains.

As AI continues to evolve, its role in logistics will become even more sophisticated, enabling businesses to navigate complex market conditions with greater agility. The integration of AI with emerging technologies such as 5G, IoT, and autonomous transportation systems will further revolutionize logistics management. By advancing AI research and fostering cross-industry collaboration, the logistics sector can achieve both economic efficiency and environmental sustainability.